\documentclass[11pt,draftcls,onecolumn]{IEEEtran}
\usepackage[normalem]{ulem}

\usepackage[normalem]{ulem}
\usepackage{cite,ctable}
\usepackage{mathtools}
\usepackage{multirow}
\usepackage{color,ctable}
\usepackage{epsfig,graphicx,amsmath,epstopdf}
\usepackage[tight]{subfigure}

\usepackage{array}

\usepackage{multirow}
\usepackage{booktabs}

\begin{document}
\title{Wavelet subspace decomposition of thermal infrared images for defect detection in artworks}

\author{M. Z. Ahmad$^{\textrm{1}}$, A. A. Khan$^{\textrm{1}}$, S. Mezghani$^{\textrm{2}}$, E. Perrin $^{\textrm{2}}$, K. Mohoubi $^{\textrm{3}}$, J. L. Bodnar$^{\textrm{3}}$, V. Vrabie$^{\textrm{2}}$ \\
Email: $13$mseemahmad@seecs.edu.pk, amir.ali@seecs.edu.pk\\
\footnotesize
$^{\textrm{1}}$School of Electrical Engineering and Computer Science (SEECS),\\ \footnotesize National University of Sciences and Technology, Islamabad, Pakistan. \\ 
\footnotesize $^{\textrm{2}}$Centr{\'e} de Recherche en STIC, \footnotesize Universit{\'e} de Reims Champagne-Ardenne, F-51000 Châlons-en-Champagne, France.\\
\footnotesize $^{\textrm{2}}$GRESPI, \footnotesize Universit{\'e} de Reims Champagne-Ardenne, F-51100 Reims, France.\\
%\footnotesize $^{\textrm{3}}$Electricit{\'e} de France (EDF), France\\ 
%\footnotesize $^{\textrm{4}}$GIPSA-Lab, Department Images and Signal, \footnotesize Grenoble Institute of Technology, France.
%\thanks{School of Electrical Engineering and Computer Sciences (SEECS), National University of Science and Technology, Islamabad, Pakistan. Email: amir.ali@seecs.edu.pk}, V. Vrabie\thanks{Centre de Recherche en STIC (CReSTIC), University of Reims, BP 1039, 51687 Reims, France.}, Y. L. Beck\thanks{Electricit{\'e} de France (EDF), $21$ Avenue de l'Europe, BP 41, 38040 Grenoble, France.}, J. I. Mars\thanks{GIPSA-Lab, Department Images and Signal, Grenoble Institute of Technology,
%BP 46, 38402 Saint Martin d'H{\`e}res, France.}, and G. D'Urso$^{\textrm{3}}$
%\thanks{
%Me and You Entreprises},
%\ Member, ASCE
%1GIPSA-Lab, Department Image and Signal Processing, Grenoble Institue of Technology,
% I've found that the \and command doesn't quite work, so just use "and"
%  such as the following (and don't forget the ending curly brace `}').
%\\
%and
%V. Vrabie%
%\thanks{Flourishing wife of same.},%
%\ Not a Member, ASCE
}

\maketitle

\begin{abstract}
%\boldmath
Monitoring the health of ancient artworks requires adequate prudence because of the sensitive nature of these materials. Classical techniques for identifying the development of faults rely on acoustic testing. These techniques, being invasive, may result in causing permanent damage to the material, especially if the material is inspected periodically. Non destructive testing has been carried out for different materials since long. In this regard, non-invasive systems were developed based on infrared thermometry principle to identify the faults in artworks. The test artwork is heated and the thermal response of the different layers is captured with the help of a thermal infrared camera. However, prolonged heating risks overheating and thus causing damage to artworks and an alternate approach is to use pseudo-random binary sequence excitations. The faults in the artwork, though, cannot be detected on the captured images, especially if their strength is weak. The weaker faults are either masked by the stronger ones, by the pictorial layer of the artwork or by the non-uniform heating. This work addresses the detection and localization of the faults through a wavelet based subspace decomposition scheme. The proposed scheme, on one hand, allows to remove the background while, on the other hand, removes the undesired high frequency noise. It is shown that the detection parameter is proportional to the diameter and the depth of the fault. A criterion is proposed to select the optimal wavelet basis along with suitable level selection for wavelet decomposition and reconstruction. The proposed approach is tested on a laboratory developed test sample with known fault locations and dimensions as well as real artworks. A comparison with a previously reported method demonstrates the efficacy of the proposed approach for fault detection in artworks.
\end{abstract}
% \fi
\IEEEpeerreviewmaketitle
\section{Introduction}
%\tcr{Section to be initiated by V.V. and completed by A.K. Start with the background on fault identification in general citing some recent references, specially from IEEE Tran. TIM or Sensors, or others.}
The detection and identification of faults in different materials has seen major emphasis over the last decades, both in research context and in industrial applications. %also in the domain of application studies, has been to detect and identify faults in several 
%The problem of non-destructive testing has been a subject of central importance for a long time in diverse industrial application fields as manufacturing, quality control, and so on as for example to characterize the presence of thin defects in conductive materials \cite{Bernieri2013}. 
Diverse techniques like Gamma and X-rays, ultrasounds, Foucault currents and Nuclear magnetic resonance are conventionally utilized for non-destructive testing of materials \cite{Baker-Jarvis1994}, \cite{Vasic2004}, \cite{Bernieri2000}, \cite{Benedetti2006}, \cite{Karagiannis2011}, \cite{Faifer2011}, \cite{Ricci2012}. 
Recently, techniques focusing on thermal radiations, such as photo-reflectance, photo-acoustics, mirage effect and photo thermal radiometry, have been tested on the laboratory scale for non-invasive testing of materials \cite{Wilson2012}. 
One such technique is the photo-thermal radiometry (or photo-thermometry) which requires a relatively simple and low-cost experimental setup \cite{Bagava2013}. 
The test material is excited by a flow of heat, resulting in a change in local thermal conditions of the material. 
The thermal response of the material to this excitation is then captured by a thermal infrared camera. 
The acquired thermal response depends on different parameters of the material such as thermal conductivity, diffusivity, emissivity and specific heat as well as the excitation used at the input. 
More specifically, the above properties manifest themselves in the thermal response depending upon different factors such as fissures or holes in the material, material structure, physicochemical processes taking place in the material, delamination, sedimentation, etc. 
The detection of faults in ancient artworks is a challenging problem requiring adequate precaution to identify the faults without deteriorating the material, even slightly.
The conventional method used for exciting the material is the pulse excitation method which consists of heating the material over a relatively long time and then allowing it to cool down \cite{Qingju2015}. 
In the current research endeavor the frescos(Italian murals) are the subject of interest which are layered structures. Due to the environmental effects air gaps occur at the interface of the layers. This is the working definition of a defect in the artwork which will be used through out the current paper.
The acquired thermal response is relatively easier to process from fault identification purposes. However, the prolonged heating involved in the process risks overheating and thus causing damage to the material. 
An alternate approach is to use random excitations like the Gaussian and pseudo-random binary sequence (PRBS) excitations. 
The step heating results in a greater transfer of thermal energy to the material under analysis as compared to the PRBS excitation.In photo-thermal radiometry the detection of fault is based on the variations in the thermal radiations based on the presence of faults hence a greater transfer of thermal energy will yield better contrast.
While these techniques are more favorable from material health conservation perspective, they necessitate more intensive processing to retrieve useful information about the defects.

The useful information should reveal the localization maps of different defects in the material after having suppressed the effects of pictorial layer, measurement noise, potential interference patterns due to multiple excitation sources and inhomogeneous illumination of the test material.
Recently, some signal processing based methods were deployed for detecting defects in artworks while using PRBS excitations \cite{vrabie2012active}. Although, the method revealed some interesting results regarding the material defects, the interpretation of the final results was not very evident. 
In this paper, we propose a new processing algorithm based on wavelet decomposition for identification of defects. 
The algorithm exploits the multi-resolution capability of wavelet transform and approximates an efficient subspace, which contains information on the defects, while removing other influences from the acquired thermal response. 
The proposed subspace decomposition approach is applied spatially and the resulting subspaces for different acquisitions are then temporally processed to extract a single representative subspace for the defects. 
Since there is no universal criterion for the selection of wavelet basis function, we propose a selection criterion as well as an automated mechanism for determining the wavelet decomposition level. 
%We also propose a method for selection of an appropriate wavelet basis function for analyzing the acquired thermal images. 
Even though the main focus of the paper is proposition of a defect detection algorithm, it is demonstrated that the intensity of the final detection parameter can be associated to the intensity  of the defect. 
%Since, the method uses random excitation, another important parameter is the duration of the random excitation pulse, i.e., the excitation time. 
%\tcb{A comparative analysis of different excitation times is performed to study their influence on the detection parameter.[Sir, shouldn't this line be removed]} 
The proposed method is validated on a sample test material, containing defects of various depths and diameters. 
The detection parameter allows localizing these faults in the test sample with different intensities. The method is also applied on a real artwork to reveal interesting information about the defects.     
The paper is organized to start with a presentation of the experimental setup, the test samples and the acquired data set. 
This is followed by the proposed methodology section comprising the mathematical formulation of the problem, the details of the proposed algorithm and the different parameters involved therein. 
The results of the proposed approach on a test sample are discussed in the subsequent section. %with an elaborate analysis on the influence of different  parameters. 
The paper is finally concluded to highlight the pros and cons of the proposed approach along with an insight into some future work in the domain.
 
%\subsection{Problem Physics}
%\IEEEpubidadjcol
%\subsubsection{Conventional NDT Techniques}
%\subsubsection{Photometry w.r.t. conventional techniques, pulse/random}
%\subsection{Need for Signal Processing }

\section{Experimental Setup and Raw Data Set}
In this section, the experimental setup for photo-radiometry system including the excitation and acquisition process is described, along with the presentation of a sample data set.  
\begin{figure}
\begin{center}
  \subfigure[Experimental setup demonstrating sample illumination and recording]{
         \includegraphics[width=7cm]{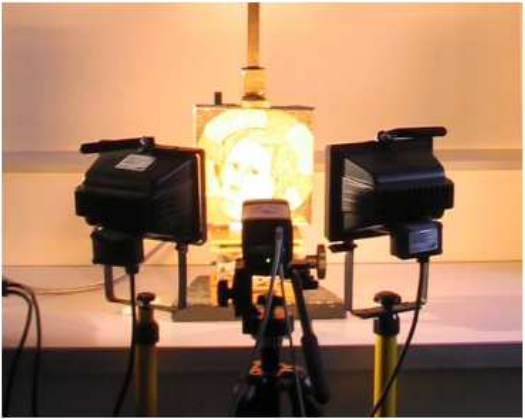}
				 \label{subfig:exp_setup}
        }
	\vspace{1cm}
	\subfigure[A typical pseudo-random excitation sequence]{
				\includegraphics[width=8cm]{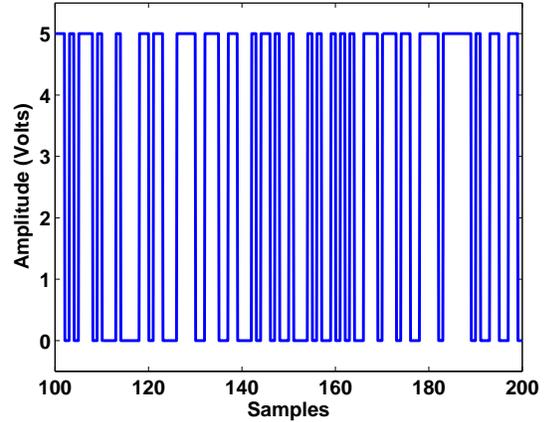}
				\label{subfig:prbs_excit}
			  }
\caption[Experimental setup for the non-invasive testing of a sample material.]{Experimental setup for the non-invasive testing of a sample material. The material is excited by two $500$ W halogen lamps, simultaneously controlled with a pseudo-random excitation sequence. The resulting thermal response of the material is captured by a thermal long wave infrared camera.} 
\label{fig:exp_setup}
\end{center}
\end{figure}
\subsection{Excitation and Acquisition System}
\label{sec:exp}
The experimental setup for photo thermometry is illustrated in Figure \ref{subfig:exp_setup}. 
The artwork is heated simultaneously by two halogen lamps, each one with a power of $500$ W. 
The lamps are symmetrically placed around the infrared camera and placed at approximately $50$ cm from the artwork.  
%%
%\tcr{It would be good to add some details regarding the power (W) and the voltage levels achievable and why they are suitable for this experimental setup, or put some relevant reference here} 
The temporal excitation pattern of these two lamps is controlled by a pseudo random binary sequence (PRBS). 
The smallest duration of each binary pulse in this sequence represents the excitation time ($T_e$). 
The experiment described in this paper is carried out for different excitation times as represented in Table \ref{tab:te}.This was done to demonstrate the generality of the proposed algorithm across different experimental parameters.
An illustrative excitation pattern is shown in Figure \ref{subfig:prbs_excit}.
 
The incident flow of heat strikes the surface of the material and its subsequent absorption depends on the physical properties of the material. 
The surface defects, such as cracks, might appear instantly due to the different optical properties and require other methods for detection.
The sub-surface defects appear after some time depending on their respective depths.
The major challenge is posed by the defects that are not visible at the surface and are located at different or varying depths.
Such defects cause air pockets in otherwise uniform material, resulting in different distributions of the heat inside the materials.
The radiated thermal energy is modified and presents an opportunity to identify the faults and to quantify them.   
In brief, different defects into a material exhibit different thermal responses to the same excitation. 

The thermal response of these different areas subjected to the excitation sequence are then captured by a A$20$ FLIR thermal Infrared (IR) camera. 
This bolometer camera, working in long wavelength range ($7.5$ to $13$ $\mu m$) and having a thermal sensitivity of $0.12{^\circ}C$ at $30{^\circ}C$, was placed around $50$ cm from the sample.
The excitation of the lamp and the IR camera were synchronized, the former being in a slave mode. 
A thermal image is obtained for each pulse duration of the input PRBS sequence, the total number of such images is thus same as the number of pulses in the PRBS sequence, $N_t$. 
The number of pulses in the PRBS sequences corresponding to the different excitation times is mentioned in Table \ref{tab:te}.

\renewcommand{\tabcolsep}{1pt}
\begin{center}
\begin{table}
 \caption{Excitation Times($T_e$) and the corresponding number of pulses in the PRBS sequences ($N_t$)}
\begin{tabular}{|c|c|c|c|c|c|}
\hline
$T_e$&$0.5s$&$1s$&$2s$&$5s$&$10s$ \\ \hline
$N_t$&$2048$&$1024$&$512$&$256$&$128$\\ \hline

\end{tabular}
\label{tab:te}
\end{table}
\end{center}
%The recorded images were of size $240 \times 320$ pixels.
%In the PRBS with $T_e=5 sec$, there are $256$ pulses and thus the same number of images are acquired along with $10$ cold images making $N_T = 266$.
%The cold images are thermal images of the artwork when no excitation is applied to the lamps and can be used to establish a baseline. 

Even though, the relative positions of the halogen lamps and the camera have been carefully setup after exhaustive testing, the problem of inhomogeneous illumination can not be ruled out. 
It will be demonstrated that the proposed wavelet based subspace decomposition approach allows estimation of a background affect which may be attributed to inhomogeneous illumination. 
The preceding description of the experimental setup indicates that the thermal behavior of the material is spatially non-stationary. 
The subsequent sections will reveal how the proposed approach allows exploiting this spatial non-stationarity.           
%\begin{figure}
%\includegraphics[width=3in]{figures/Prbs_seq_1024_100_200.eps}
%\caption[A Sample Pseudo-Random Excitation Sequence.]{A pseudo-random excitation sequence (PRBS) %controls the excitation of both the lamps simultaneously.Each sample of the PRBS is of uniform duration in time. This duration is called the Excitation Time. The results vary with this parameter as we will discuss in later sections.}
%\label{fig:prbs_excit}
%\end{figure}

\subsection{Test Materials}
Before developing a mathematical formulation for the problem of defect detection, we briefly shift our focus to the presentation of the test samples used in this work. In particular, we consider two types of test materials, one with faults of different dimensions simulated by introducing holes in a plaster material and other are real artworks containing faults of varying geometry and dimensions.

\subsubsection{Laboratory test sample }
\begin{figure}
  \begin{center}
  \subfigure[Front side directly exposed to excitation]
         {\includegraphics[width=7cm]{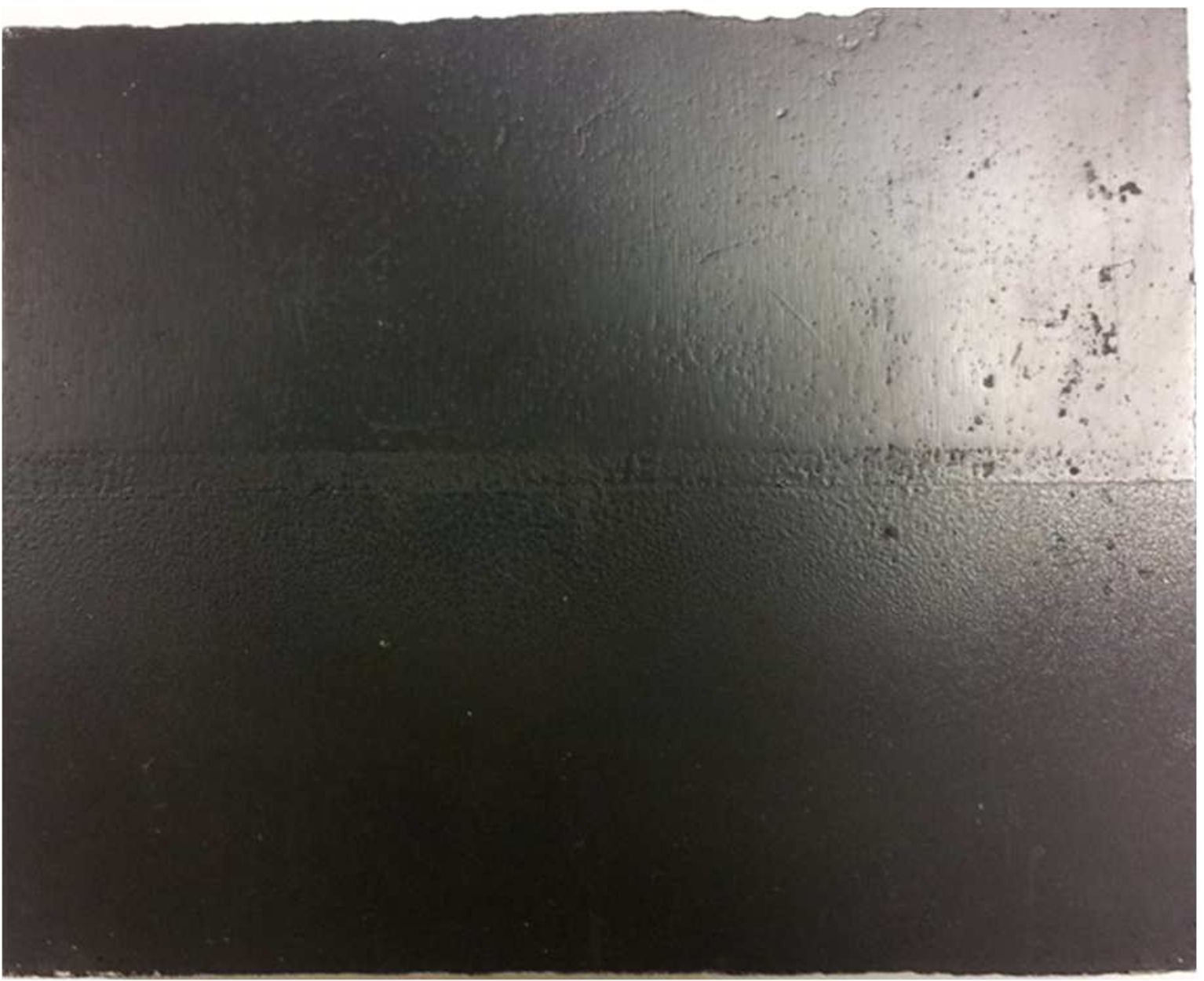}
         \label{test_sample_a}}
  \subfigure[Back side containing holes to simulate defects]
        {\includegraphics[width=7cm]{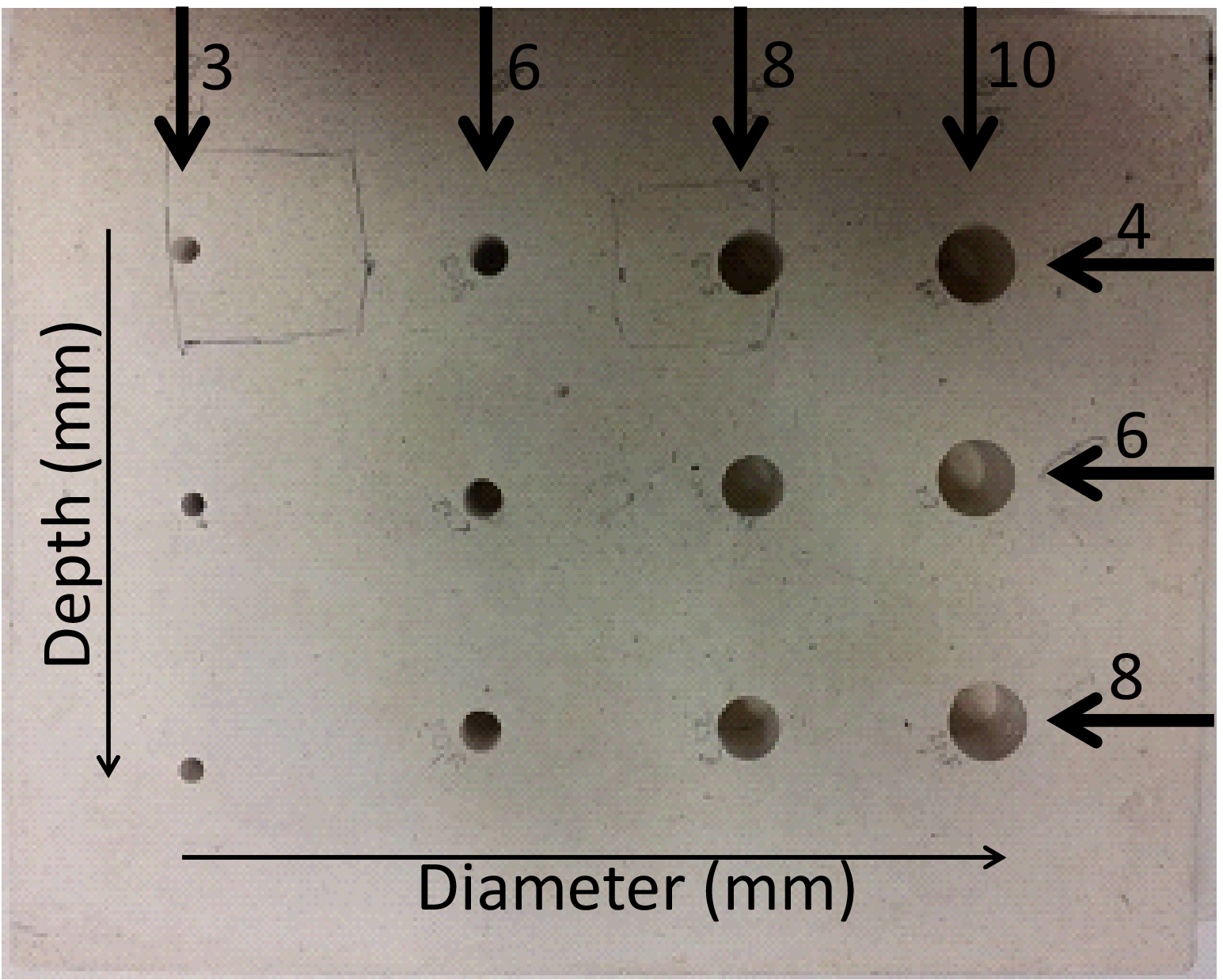}
         \label{test_sample_b}}
   \end{center} 
  \caption[Synthetic test sample prepared in laboratory containing holes.]{Front and back sides of the test material prepared in laboratory. The material is constructed from plaster with the front side exposed to the flow of heat containing no holes or other significant anomalies. The back side of the material consists of $12$ holes of varying depths and diameter to simulate defects. Depths are indicated from the front side.} 
  \label{fig:test_sample}
\end{figure}

The complete acquisition chain of the experimental setup will be demonstrated with the help of this test material to enhance understanding of the underlying problem, which needs to be modeled. 
It will also serve as a ground truth against which different parameters of the proposed algorithm will be established. 
The test material is composed of a white plaster substrate with a thickness of approximately $20$mm (Fig. \ref{fig:test_sample}).
The front side of the material, which is exposed to the excitation, does not contain any significant defects, but still has some minor irregularities, especially toward the edges (Fig. \ref{test_sample_a}). 
%%%%%%%%%%%%%%%%%%%%%%%%%%%%%%%%%%%%%%%%%%%%%%%%%%%%%%%%%%%%
%%%%%%%%% 
%\tcr{DEPTH OF THE FAULTS ARE FROM THE FRONT SIZE. YOU SHOULD ROTATE IMAGES and MODIFY THE TABLE. CHECK MODIFIED EXPLICATIONS }
As described earlier a defect in an artwork is typically characterized by the slight delamination of the material in the vicinity of the defect causing air pockets within the material. 
The back side of the material was therefore modified to create defects of different dimensions (Fig. \ref{test_sample_b}). 
In particular, differentiation amongst these defects arise from differences in their depths and diameters. 
Twelve different holes were created in the material and their dimensions (depths and diameters) are marked in Fig. \ref{test_sample_b} and also summarized in Table \ref{tab:holeinfo}. 
It is pertinent to mention that these depths are measured from the front side of the material. 
Thus, hole $1$ (Fig. \ref{test_sample_b} top right) is at a depth of $4$ mm from the front surface. 
% Reverifier et modifier tableau \label{tab:holeinfo}
The holes have been numbered in sequence starting from top right (hole $1$) and moving left and down to bottom left (hole $12$). 
As illustrated in Fig. \ref{test_sample_b}, three different depths, ($4$ mm, $6$ mm, and $8$ mm) with four different diameters ($3$ mm, $6$ mm, $8$ mm, and $10$ mm) for each of these depths, have been tested in this experiment. 
It is assumed that each fault has uniform depth and diameter signifying that at the spatial location of the fault, there's an air pocket across the material up till the depth of the given fault.
As the surface is heated the thermal energy gradually diffuses through the depth of the material. The rate of the diffusion depends on the thermal diffusivity of the material. The thermal diffusivity of the plaster of Paris (the main constituent of the fresco) is greater than that of the air (present at the fault locations). This results in a difference in the rate of diffusion of heat and in result the amount of back scattered radiation results in detection of fault regions(as they present higher thermal intensity).
Shallower and wider faults offer bigger air pockets which effect the thermal diffusivity and conductivity.
This translates physically into the $4$ mm hole (depth) more likely to depict a higher heat intensity as compared to the $6$ mm and $8$ mm holes. 
%The different holes have been labeled as in Table \ref{tab:holeinfo} for ease of reference during the subsequent analysis.
%The hole on the top right is numbered $1$, hole on top left is numbered $4$ and in similar progression the bottom left hole is numbered $12$. 
%Figure Fig. \ref{test_sample_b} illustrates the holes with three different depths ($8$ m, $6$ m and $4$ m). 
%Likewise, four different diameters ($10$ m, $8$ m, $6$ m and $3$ m) have been tested for each of these depths. 
%The depth of the hole is being measured from the back surface of the material thus a hole with greater depth is effectively closer to the front surface (shallower) and vice versa.
%It is assumed in this control test that each fault has uniform depth and diameter and that fault does not occur in the center of the material rather at the back surface.
%This means that at the location where the fault occurs there is an air pocket throughout the material up till the depth of the given fault. 
%This affect implies that deeper and bigger (w.r.t. diameter) holes offer bigger air pockets which effect the thermal diffusion and conductive properties. 
The overall impact on the outcome is that the acquired thermal response is more intense at the location of shallower and wider holes as compared to deeper and narrower holes.

\renewcommand{\tabcolsep}{1pt}
\begin{center}
	\begin{table}
\caption{Characterization of laboratory prepared test sample (Fig. \ref{fig:test_sample}): co-ordinate mapping of the faults' (holes') centres; dimension of the faults; and the improvement in the SNR (difference between the SNR after the application of proposed algorithm and the SNR of the Raw Data) after the application of proposed algorithm calculated as in Equation \eqref{eq:SNR}.}
    	\begin{tabular}{|l| c| c| l| l| l| l|l|l|l|l|}
        \hline
     \textbf{Hole} &\multicolumn{2}{|c|}{\textbf{Coordinates(pixels)}}& \textbf{Diameter}& \textbf{Depth}&\multicolumn{5}{|c|} {\textbf{Improvement in SNR via proposed algorithm}} \\ \hline
    %\cmidrule(r){2-3}
     & $\hspace{2pt} N_x \hspace{2pt} $ & $N_y$ &(mm) & (mm)  & \multicolumn{5}{|c|}{(dB)}\\ \hline
		& & & & &  $Te=0.5s$&$Te=1s$&$Te=2s$&$Te=5s$&$Te=10s$ \\ \hline
    %\midrule
    $1 $      &  $39   $  &  $167 $  &    $10  $   &  $ 4$ &$9.9579$&$8.0364$& $7.5142$&$9.7085$& $9.8548$ \\ \hline
    $2 $      &  $39   $  &  $127 $  &    $8    $   &  $ 4$&$4.5801$&$3.8630$& $5.2932$&$9.2458$ & $3.5419$ \\ \hline
    $3 $      &  $42   $  &  $83   $  &    $6    $   &  $4 $&$-1.5678$&$-2.8933$& $2.4168$&$5.5109$& $2.2004$\\ \hline
    $4 $      &  $41   $  &  $30   $  &    $3    $   &  $4 $ &$9.3600$&$7.9558$& $ 7.3496$&$-1.0940$& $10.7252$\\ \hline
    $5 $      &  $77   $  &  $167 $  &    $ 10   $   &  $6$ &$4.9934$&$4.5480$& $3.7430$&$6.2764$& $6.1833$\\ \hline
    $6 $      &  $ 76  $  &  $127 $  &    $8  $   &  $6 $ &$2.2338$&$-2.6496$& $2.8383$&$5.4974$  & $6.4358$\\ \hline
    $7 $      &  $83   $  &  $83   $  &    $6    $   &  $6 $ &$4.8059$&$5.3750$& $5.4407$&$2.1080$ &  $6.2786$\\ \hline 
    $8 $      &  $86   $  &  $31   $  &    $3    $   &  $6 $ &$-5.5033$&$2.9049$& $2.5581$&$1.0447$ &  $-0.1188$\\ \hline
    $9 $      &  $117 $  &  $169 $  &    $ 10  $   &  $8 $ &$-5.1488$&$-2.4098$& $-2.8700$&$2.4369$  &  $1.4146$\\ \hline
    $10$      &  $118 $  &  $127 $  &    $ 8   $   &  $8 $ &$-2.1426$&$1.6670$& $-2.9004$&$5.2986$   & $-5.9618$\\ \hline
    $11$      &  $122 $  &  $83   $  &    $6  $   &  $8 $ &$2.6452$&$-2.6246$& $0.4480$&$-2.2460$    & $-1.4110$\\ \hline
    $12$       &  $130 $  &  $31   $  &    $3    $   &  $8 $ &$3.8080$&$-0.5390$& $-4.0005$& $-2.5897$& $5.8405$ \\ \hline
  
    \end{tabular}
    
    \label{tab:holeinfo}
 \end{table}
\end{center}

Photo-radiometry is an inspection technique, which works at depths closer to the contact surface \cite{Bagava2013}. 
While, the higher intensity faults may be identifiable in the raw thermal images, the smaller faults are often masked. 
The goal of this work is to extract such faults from the background. 
The test sample described above will also allow studying the influence of the diameter (spread) of a certain defect for a given depth. Moreover, it will reveal the impact of processing raw data through the proposed algorithm in terms of enhancing the detectability of the faults at different depths. 
%and an objective of this research is to establish these depth limits for the defects. 
%Else where, this study will  
It should be mentioned that real defects in an artwork will not necessarily be circular in shape, but the purpose of the current research is not to indulge in particular defect geometries but rather to establish the detectability limits. 
Nevertheless, the viability of the proposed approach for irregular defects will be demonstrated on real artworks in the results section. 

\subsubsection{Real Artwork}
\label{subsubsec:realtestdata}
\begin{figure}
	\begin{center}
	\subfigure[Picture of Mural $1$]
			{\includegraphics[height=6.75cm,width=7cm]{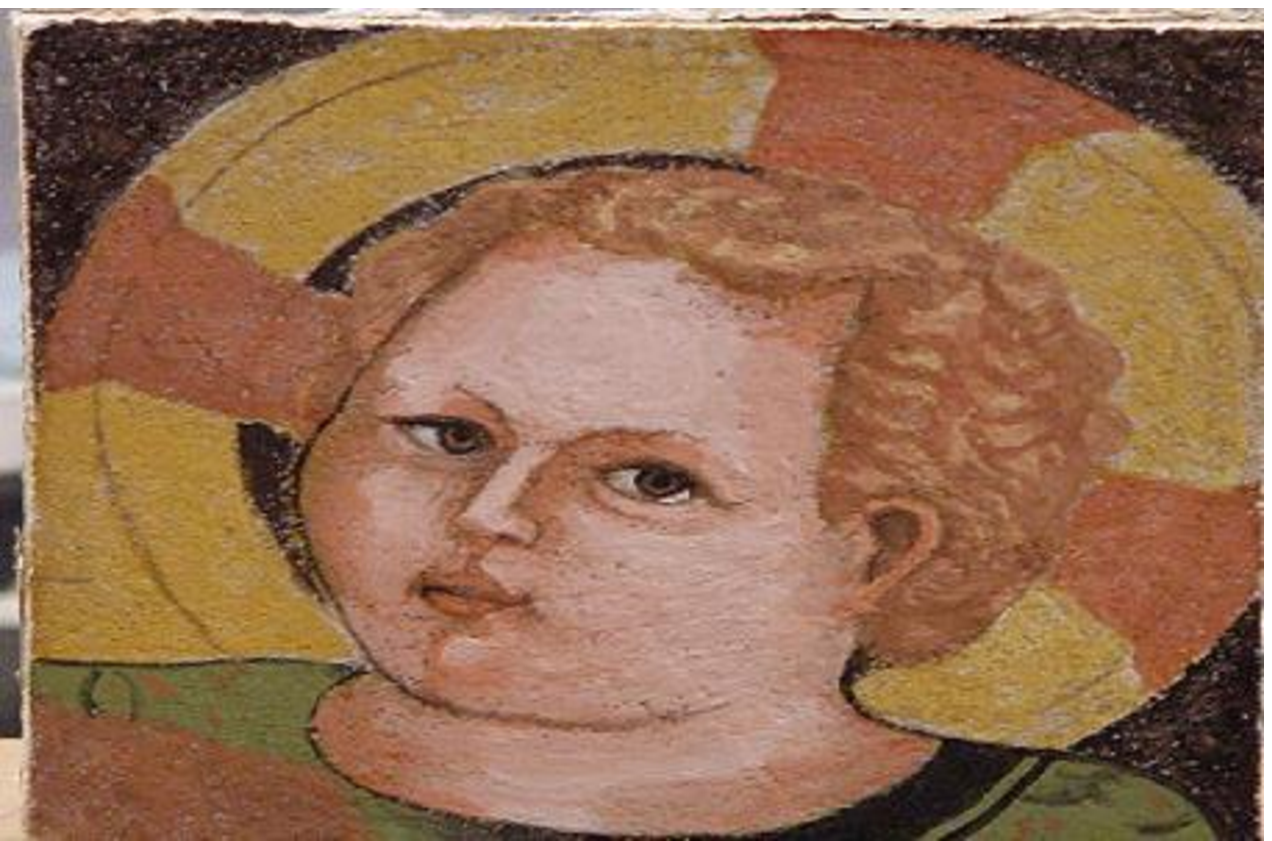}
			\label{stchris}}
	\subfigure[Fault map of Mural $1$]
			{\includegraphics[width=7cm]{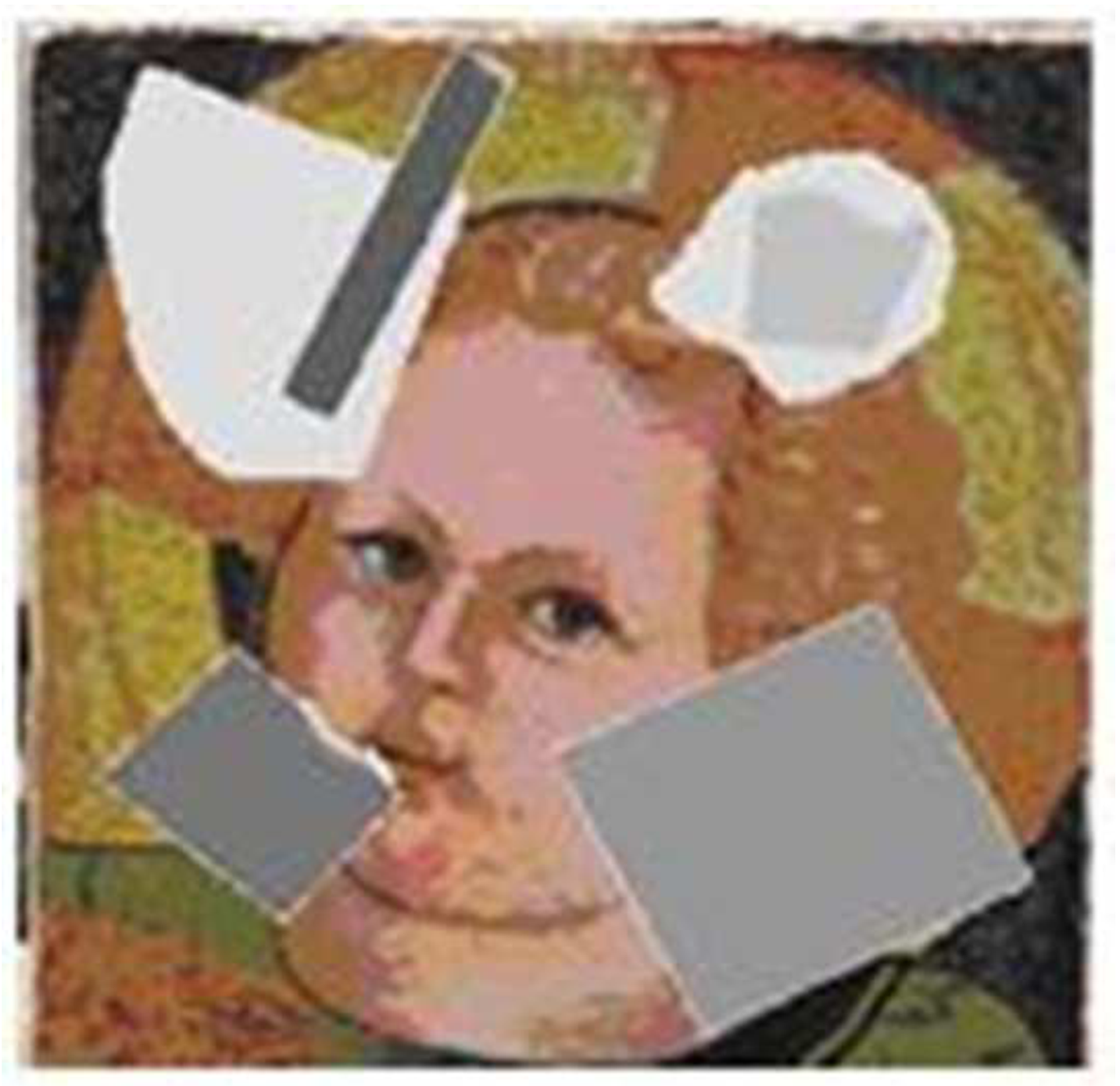}
			\label{stchris_fault}}
	\end{center}
	
	\caption[St. Christophe photograph and fault map]{A real artwork, Mural $1$: (a) photograph of Mural $1$ displaying no visible defects; (b) fault map highlighting the locations of faults and their irregular geometry for performance analysis. The faults $\bf{A}$ to $\bf{E}$ are incorporated with different dimensions.}
	\label{fig:test_sample_stchris}
	\end{figure}

In order to validate the proposed approach, thermal responses of a real artwork, namely Mural $1$, were acquired.
This artwork represents a replica of the ``Saint Christopher with the Christ Child'', a Florentine fresco realized in the end of the $14^{th}$ century, created in laboratory according to the technique of the Italian primitives ~\cite{vrabie2012active}.
Fig. \ref{stchris} shows a picture of Mural $1$, along with the designated faults in this artwork (Fig. \ref{stchris_fault}). 
Five inclusions of plastazote have been introduced in the fresco in its manufacturing process. 
These five faults are labeled \textbf{A} to \textbf{E}: \textbf{A} is tilted with varying depths of $3$ to $10$ mm and a thickness of $5$ mm; 
\textbf{B} is at a depth of $5$ mm with a thickness varying between $3$ mm and $10$ mm; 
\textbf{D} is located at a depth of $3$ mm and has a thickness of $3$ mm; 
\textbf{E} is at a depth of $10$ mm, lying beneath \textbf{D}, and has a thickness of $5$ mm; 
finally, \textbf{C} is a powdery defect located at a depth of $3$ mm and has thickness of $5$ mm. These faults are of non-uniform shapes.

 Another real artwork used during the course of this work is named Mural $2$. No information regarding the position of the sub-surface faults is known prior to the application of the proposed algorithm. The surface artifact is a piece of gold foil which in the course of present investigation is not of interest and will hence be removed using post processing techniques as discussed later on.
 
\section{Proposed methodology: selective sub-space extraction for defect detection}
\begin{figure}
	\begin{center}
    \includegraphics[width=9cm]{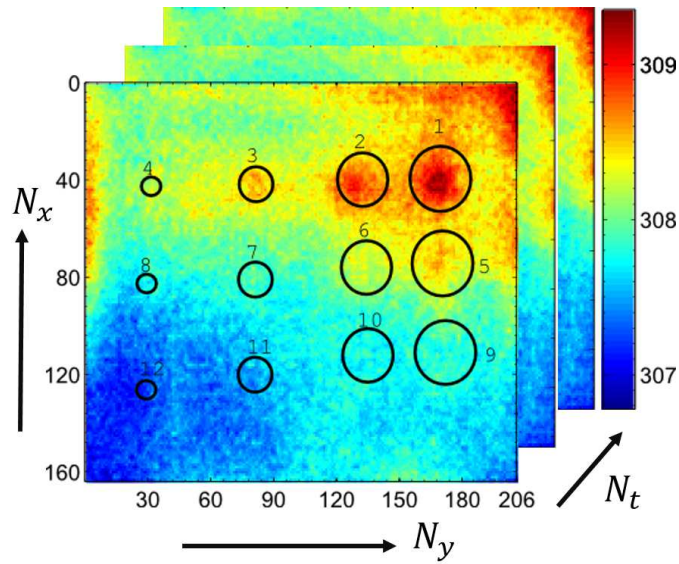}
    \caption{Raw data $\mathcal{Y}$ acquired on the laboratory generated test material of Fig. \ref{fig:test_sample}. One thermal image per acquisition pulse is acquired thus generating a data cube. The colors represent the thermal intensities as measured by the camera.}
    \label{fig:rawdatacube}
  \end{center}
	\end{figure}

In this section, we will present the proposed approach for the detection of defects. Before delving into the details of the algorithm, we will associate a mathematical formulation to the acquired thermal responses. Moreover, we will also briefly discuss a method previously developed by the authors in order to substantiate the merits of the approach proposed in this paper.
The acquired raw data consist of $N_{t}$ thermal images, as shown in Fig. \ref{fig:rawdatacube}. Note that images acquired on the front side were flipped on the $y$ axis in order to have the same positioning of the holes shown in the back side (see Fig. \ref{test_sample_b}).
The raw data cube, $\mathcal{Y}$, exist in the Hilbert space $\mathcal{Y}\in {\Re}^{N_{x} \times N_{x} \times N_{t}}$, where $N_{x}$ and $N_{y}$ correspond to the dimension of the test material (pixels) as captured by the camera and $N_{t}$ corresponds to the total number of acquisitions in time (samples). 
While $N_{x}$ and $N_{y}$ will be fixed for a given test sample (assuming the experimental setup is not changed and the resolution of the camera is not exceeded), the value of $N_{t}$ will depend on the excitation time ($T_e$) as represented in the Table \ref{tab:te}. 
%The influence of this parameter will be discussed in the results section. 
The raw data contains information from multiple sources including defects, background or pictorial layer, inhomogeneous illumination and measurement noise. 
The Hilbert space containing the raw data can be decomposed into its subspaces, each representing single or multiple afore mentioned sources. 
Subspace decomposition is a processing technique which allows more meaningful projections. In our case, we are looking to decompose the data into background, defect (useful) and noise subspaces.
Subspace decomposition based matrix filtering techniques have been used in diverse applications \cite{Zhang2001a}, \cite{Ece2004}, \cite{Blanco-Velasco2008}, \cite{Cusido2008}, \cite{khan2008source}. 
Singular Value Decomposition (SVD) and Wavelet decomposition are common methods of sub-space decomposition.
In this paper, the SVD based technique for defect identification will be compared with the proposed wavelet based decomposition technique.

\subsection{State-of-the-art decomposition of thermal responses}
\label{subsec:SVDweakness}
An SVD based defect detection scheme in artworks using the thermal responses was developed by Vrabie et al. \cite{vrabie2012active}. 
The scheme was based on the hypothesis that the evolution of thermal response over time is a distinguishing factor between defects and background. It involved rearranging the data cube, $\mathcal{Y} \in {\Re}^{N_{x} \times N_{x} \times N_{t}}$ into a $2$-D data set ${\bf Y} \in {\Re}^{N_{s} \times N_{t}}$, where $N_{s} = N_{x} \times N_{y}$, represents the spatial dimension. 
This $2$-D space was then decomposed into three subspaces, the first one representing background, the second one useful information containing defects and the third one the noise components. 
The number of singular values to be considered for each of the three subspaces was determined empirically. 
Specifically, only one singular value containing $96\%$ of energy was used for extracting the background, whereas the so called useful subspace containing defects was synthesized from few subsequent components and the lowest energy singular components were attributed to the noise. 
The second subspace was selected for further processing to detect the locations of defects. 
The higher order statistics (HOS) of skewness and kurtosis were then computed for the time series in the useful subspace to obtain the results in ${\Re}^{N_{s}}$ space, which were then rearranged to obtain an image map in ${\Re}^{N_{x} \times N_{y}}$ for both kurtosis and skewness. 
It was demonstrated that these skewness and kurtosis maps can help in identification of defects. 
However, the interpretation of the results obtained with this SVD and HOS based approach was not systematic. 

Although, SVD is a matrix filtering technique, it was utilized in the temporal domain without any focus on explicitly exploiting the spatial patterns of the defects. 
The high energy first component of SVD contains information on the average behavior of the input data. 
Since, the SVD-HOS approach, discussed above, discards the first subspace, it therefore runs the risks of suppressing high energy defects along with the background. 
This fact is attributable to the poor separability of different components by a second-order statistics based approach such as SVD. 
The limitations of this SVD-HOS based approach led us to the development of an alternate approach for detection of defects. 
This multi-resolution analysis based approach also performs the subspace decomposition but in spatial rather than the temporal domain. 
%It assumes that the fault information, background and the noise will appear in different components and thus get separated.We treat each pixel, individually, as a time series.Then we apply Higher Order Statistics(HOS) on the decomposed sub-space of our interest to determine whether the pixel belonged to the fault or the background.In the wavelet decomposition we consider an image at a particular time-stamp and then apply the statistical measures on the useful sub-space.\newline
%The method used for the sub-space decomposition is only the pre-processing stage. The later stages including the application of statistics holds for both the methods.SVD is adversely effected by the presence of a trend. In the SVD the trend of the data appears as the strongest component.We assume that the information of the pictorial layer the most abundant will manifest in this component and the noise in the higher components. The components in between will contain the information of fault locations. We can not determine which components will contain the faults and vice versa. In case of wavelet decomposition we are decomposing in the frequency domain and are trying to locate and characterize the faults w.r.t depth and diameter.     

\subsection{Wavelet based subspace decomposition}
The raw thermal response acquired by the camera at a given instant, ${\bf Y}_{\textrm{t}}\in {\Re}^{N_{x}\times N_{y}}$, incorporates information on the defects but this information is often masked by many undesirable effects such as inhomogeneous illumination, background pictorial layer and measurement noise. 
The inhomogeneous illumination is caused by irradiation from two different sources whose placement, though carefully adjusted, could not make the captured image illumination invariant. 
The characteristics of the pictorial layer of the test material such as pattern and/or the paint also manifest themselves strongly in the acquired raw image. 
In addition, the defects can have different dimensions in terms of the wear-ability of the material. 
The stronger defects will have more significant intensities and will appear sooner as compared to the weaker ones because of the fact that for stronger defects (e.g. those closer to the surface), the heat diffusion will be greater as compared to weaker ones. 
The spatial pattern of the raw thermal image is therefore non-stationary. 
In order to better exploit this non-stationarity, we resort to the joint-time frequency analysis of wavelet transform. 
The underlying principle is that of the multi-resolution analysis whereby it is possible to decompose the input data into multiple subbands. 

\begin{figure}
	\begin{center}
	\includegraphics[width=9cm]{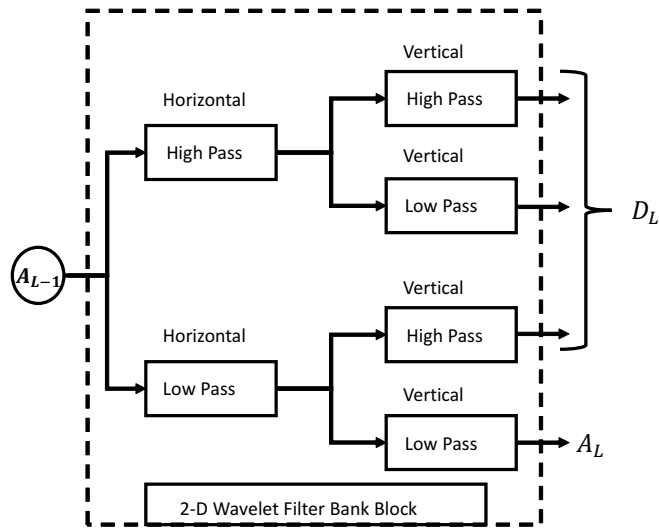}
	\caption[$2-D$ Wavelet Filter Bank]{$2-D$ Wavelet decomposition at level $L$. The low pass image ${A_{L-1}}$ at a given time instant is first horizontally filtered through $1$-D low pass and high pass filters followed by a vertical pass on the results. $D_L$ corresponds to the three details and $A_L$ represents the approximation space.}
	\label{fig:dwt}
	\end{center}
	\end{figure}

Discrete Wavelet Transform (DWT) is a powerful tool for performing multi-resolution analysis and has been exploited in numerous fault detection applications \cite{Cusido2008}, \cite{Purushotham2005}, \cite{Youssef2003}, \cite{Zhang2001}, \cite{Liu2008}, \cite{Ece2004}. 
At any level of decomposition `$L$', DWT decomposes the input data into a high pass or detail subspace $D_{L}$ and a low pass or approximation subspace $A_{L}$, with input to the level $L=1$ being ${\bf Y}_{\textrm{t}}$ and to any subsequent level $L$, the approximation of the previous level $A_{L-1}$. 
To apply the DWT on $2$-D datasets, the filtering is first applied along one dimension and then the resulting images are processed along the other dimension. $D_{L}$ represents the detail components resulting from high pass filtering in at least one dimension (horizontal or vertical) whereas $A_{L}$ corresponds to the low pass filtering output in both dimensions. 
As depicted in Fig. \ref{fig:dwt}, application of DWT on images yields $4$ sub-images, out of which the sub-image resulting from the two low-pass filters represent the approximation subspace and the remaining images are summed up to get the detail subspace. 
The output of the wavelet decomposition process depends on the basis function used for computation. 
In general, the basis function should be similar to the features being searched for in the data, however, the exact signatures of the features being not known, this choice is mostly empirical. 
In this paper, we perform an exhaustive search approach to select the best basis for our defect detection application which will be discussed later. 
Another important parameter is the number of levels to be used for decomposition which will mainly depend on the separability between the strongest defects, the background pictorial layer and the illumination effects. 
We utilize a criterion based on regional mutual information \cite{russakoff2004image} for the selection of number of  decomposition levels. 

\subsubsection{Proposed algorithm}
\begin{figure}
\begin{center}
\includegraphics[height=1.5in]{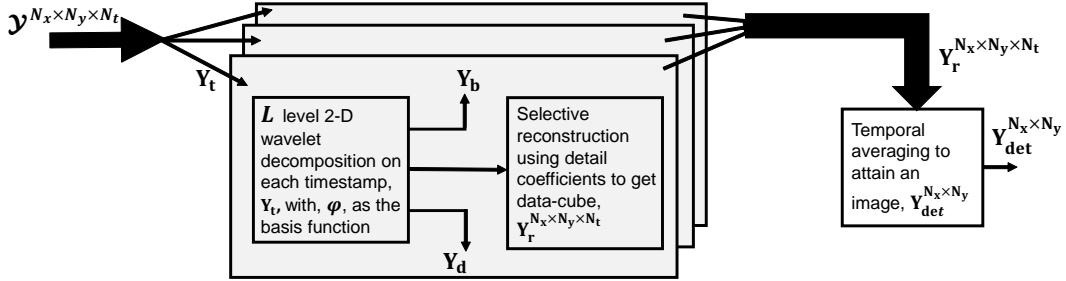}
\caption[Block Diagram of the proposed Algorithm]{The proposed scheme based on wavelet decomposition and selective reconstruction. The main parameters of the scheme are the wavelet basis $\phi$ and the number of decomposition levels, $L$. $\bf{Y_r}$ is reconstructed from intermediate level details and contains information about the faults.}
\label{fig:blk}
\end{center}
\end{figure}

The acquired data at a given instant, ${\bf Y}_{\textrm{t}}$, being a mixture of different sources (inhomogeneous illumination, measurement  noise, pictorial layer, background effects and useful defects), the goal is to separate out the useful band pass region containing the information of the defects. 
DWT with its multi-resolution capability acts as a filter bank with pass bands of different widths. 
Since our objective is the localization of fault in the spatial plane, we apply the $2$D wavelet transform in the spatial domain at each time instant. 
The proposed algorithm is presented in Fig. \ref{fig:blk} and involves three principal stages: multi-level decomposition of each image, selective reconstruction and temporal averaging to obtain the final detection result in terms of a fault map (image).    
The relevance of the DWT output depends on the choice of the basis function. 
The signal representation using a basis set essentially corresponds to finding the inner product so to obtain the best representation, the basis function must be similar to the analyzed signal. 
The acquired data, ${\bf Y}_{\textrm{t}}$, %corresponds to the thermal response of the test material to active heating by the halogen lamps which themselved are controlled by a certain random binary sequence. $\bf{Y}$ 
is dependent on the excitation sequence, the thermal properties of the test material and the nature of existing faults. 
As the nature of the faults is not known beforehand and we are not physically quantifying the material, we propose a method for determining the best basis function for our application based on an exhaustive search approach.
  
After decomposition using selected basis, the useful information containing the defects is extracted by reconstruction using appropriate subspaces. 
The space containing the raw data ${\bf Y}_{\textrm{t}}$ is decomposed into three subspaces as: 
\begin{equation}
{\bf Y}_{\textrm{t}} = \bf Y_{b}+Y_{r}+Y_{d},
\end{equation}
where $\bf{Y_{b}}$ represents the subspace corresponding to the background pictorial layer and uneven illumination effects as obtained in the approximation $A_L$ at level $L$, $\bf{Y_{r}}$ corresponds to the useful subspace containing fault information and is reconstructed from intermediate level details and $\bf{Y_{d}}$ represents the discarded subspace corresponding to the high frequency noise appearing in the lower level details.  

Three important parameters in the proposed algorithm (Fig. \ref{fig:blk}) are the wavelet basis selection; the number of decomposition levels and; the selection of the details to reconstruct $\bf{Y_{r}}$. Before focusing on the selection criteria for these three parameters, we consider the third module of the overall processing algorithm (Fig. \ref{fig:blk}). 
The first two modules of image decomposition and reconstruction are applied to every temporal acquisition thus resulting in a representative fault subspace at every time instant. 
Since the final goal is to localize the faults in the spatial domain while using PRBS excitations, the fault subspaces $\bf{Y_{r}}$, at different time instants, are temporally averaged to obtain the final result $\bf{Y_{det}} \in {\Re}^{N_{x} \times N_{y}}$. $\bf{Y_{det}}$ thus represents the spatial fault map of the sample being tested.

\subsubsection{Parameter selection criteria}
\label{subsec:Paramopt}
%\subsubsubsection{Wavelet basis selection}
An important aspect in the application of wavelet transform is the choice of basis function to be used in the decomposition and reconstruction of the subspaces. 
Although, wavelet analysis has found widespread applications in diverse domains, it is not a data driven technique and there does not exist a global rule to select the best basis. 
The most frequent approach is to perform exhaustive testing with different basis functions and select the one which produces the best desired results.
As the current problem is based on the classification of thermal image pixels into fault and non-fault regions we require a ground truth to establish the effect of the values of different parameters on the classification accuracy of the proposed algorithm. The test sample created in the laboratory represented in the Figure \ref{test_sample_b}, presents an excellent test subject as it simulates faults of differnt depths and diameters. 
For our application, we propose a criterion for best basis selection by formulating and subsequently minimizing an error function. 
We exploit the detection results on the test sample (Fig. \ref{fig:test_sample}) to select the basis which provides best separation of the desired faults from the background without compromising on the weaker faults. 
The cost function is formulated as:
 \begin{equation}
\min\limits_{j\in{\mathbf{\Phi}}}\left [ \frac{\sum\limits_{i\geq 2}{\|{\bf{Y_{det}}(win)}_{1,j}-{\bf{Y_{det}}(win)}_{i,j}\|}}{\|{\bf{Y_{det}}(win)}_{1,j}-{\bf B}_{avg,j}\|} \right ],
 \label{eq:cstfn}
 \end{equation}
where, ${\mathbf{\Phi}}$ represents the set of all the basis functions $\phi_{j}$, which are tested, ${\bf{Y_{det}}(win)}_{i,j}$ represents the detection parameter at the output of the algorithm computed in a window around the $i^{th}$ fault for the $j^{th}$ basis, ${\bf B}_{avg,j}$ represents the average value of the detection parameter in the background, where there are no known faults for the $j^{th}$ basis and ${\|.\|}$ represents the matrix norm. 
%For the ease of representation introduce a new notation. We use $a$ to denote the window size, $\Phi$ to denote the basis vector set,$\phi_{j}$ to denote the $j^{th}$ basis vector. $F_{i,j}$ is used to denote the window with the $i^{th}$ fault pixel, as shown in table \ref{tab:holeinfo},at the center and $B_{k,j}$ to denote the window with the $k^{th}$ background pixel, as shown in table \ref{tab:noiseinfo}, at the center. The subscript $j$ denotes the vector basis used for wavelet decomposition. An important thing to note is that both the $F_{i,j}$ and $B_{k,j}$ are square matrices of size $a$.
% \begin{equation}
%B_{avg,j}=\frac{\sum_{k}{B_{k,j}}}{k}
% \end{equation}
% \begin{equation}
%\min\limits_{j\in{\mathbf{\Phi}}}\left [ \frac{\sum\limits_{i=1}^{12}{\|F_{1,j}-F{i,j}\|}}{\|F_{1,j}-B_{avg,j}\|} \right ]
% \label{eq:cstfn}
% \end{equation}
%\begin{equation}
%\|A\|=\left [\sum\limits_{i=1}^{a}\sum\limits_{j=1}^{a}(a_{ij})^2 \right]^{\frac{1}{2}}
%\label{eq:frob}
%\end{equation}
The numerator of Eq.~\ref{eq:cstfn} ensures that the difference between the detection results for all the faults is minimized with respect to the most significant fault that corresponds to Hole $1$, whereas the denominator maximizes the differences between the most significant Hole $1$ and the background that corresponds to the noise. 
In fact, viewing the problem in terms of the classification of raw image into defective and non-defective regions, the minimization of this cost function ensures that the inter-class distance is maximized while simultaneously minimizing the intra-class distance. 
The best basis $\phi_{j}$ is the one that provides minimum value for this cost function. %In the equation \ref{eq:cstfn} numerator is the measure of intra-class distance and the denominator is the measure of the inter-class distance. 
%Minimizing this equation will result in either the minimizing of intra-class distance, the maximizing of the inter-class distance or both depending on the data and the basis vectors used.

It is pertinent to mention that the proposed approach has been formulated given a ground truth about the nature (depth and diameter) and the location of the faults. 
The results obtained are directly applied to the real artworks in this work with the assumption that the test sample contains a wide range of faults from very stronger ones to very weaker ones. 
Nevertheless, the measure can be improved by rigorous training on simulated faults of different shapes, sizes, depth and orientations which goes beyond the scope of current work.

%\subsubsubsection{Level Selection for Wavelet Decomposition}
The second important parameter in the proposed algorithm is the number of levels till which the original space should be decomposed with the selected basis function.
Wavelet decomposition level, when viewed in terms of filter banks, determines the frequency resolution of the proposed method and will thus depend on the nature of the defects that need to identified. 
Although, multiple factors may influence this level selection including the geometry and physical properties of the material, the two fundamental parameters that are considered for level selection are the depths and the diameters of the defects. 
We exploit the laboratory developed test sample (Fig. \ref{fig:test_sample}) to establish a criterion for level selection by computing mutual information based similarity index between the original image and its subsequent low pass versions at different levels. %This parameter must be automatically optimized by the method itself so as to eliminate the need of the prior knowledge of faults so that it can be deployed independently.
%We performed some experiments on the test data shown in Figure~\ref{fig:rawdatacube},to devise a method of automatically selecting the wavelet decomposition level. 
%For this purpose the relationship between the wavelet decomposition and the nature of the fault must be known. 
%o achieve this purpose we are using an assumption which is explained in the following paragraph.

The wavelet decomposition yields higher frequency information of the analyzed content in lower decomposition levels and lower frequency information in the higher decomposition levels. 
The faults that are closest to the surface and widest manifest themselves with the highest intensities in the acquired thermal images vis-a-vis other faults of smaller dimensions that are farther from the surface. 
As a result, stronger the fault in terms of its diameter and depth, the more likely it is to appear in the lower frequency bands of the spectrum and thus in the lower decomposition levels. 
Based on this assumption, while the smaller and weaker faults may be retrieved in some latest levels of decomposition, the stronger faults are likely to be associated with the background in the subsequent levels.  
%Our method for the wavelet decomposition level selection assumes that the stronger fault will always cover a greater frequency region than the weaker faults and thus will be present in the higher level of decomposition.
The wavelet decomposition essentially involves successive application of two-channel (low pass and high pass) filtering of the approximation (low pass content) at each level with perfect reconstruction capability. 
Hence, when at a certain level $L$, information about the strongest fault has been completely detached from the approximation at that level, $A_L$, the remaining fault information essentially goes into the detail $D_L$ at that level. 
Any further decomposition performed after this level $L$ would thus yield no additional information regarding the faults. 
We utilize a measure based on the mutual information for automating the level selection. 
%$D_{1,\cdots,L}$
%This requires visual observation of the approximation at each decomposition level and thus prevents the automation of the level selection procedure.
%To overcome this hurdle we use the concept of mutual information (MI).

Mutual information between two random variables allows measuring the similarity between them and the mutual information based measures have been extensively used as similarity index in image processing domain \cite{Pluim2003}, \cite{Crum2014}, \cite{Hirschmuller2008}, \cite{Peng2005}, \cite{corsini2009image}, \cite{Gueguen2014}. 
A slight variant of the classical mutual information is the Regional Mutual Information (RMI) \cite{russakoff2004image}. RMI exploits localized regional information rather than the global information and has been demonstrated to be a more robust method for measuring the mutual information between two images. A higher RMI indicates a higher degree of dependence between the analyzed images. 

\begin{figure}
\begin{center}
\includegraphics[width=3in]{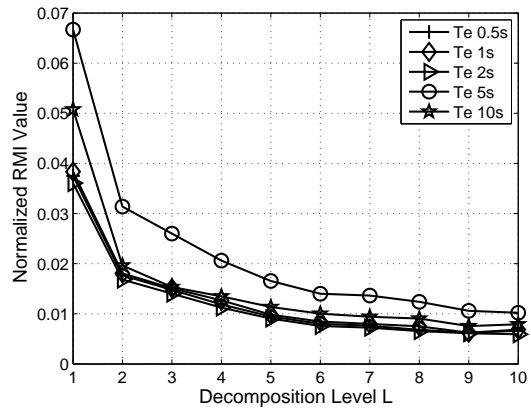}
\caption[Mutual Information of the wavelet basis db10 for different excitation times ($T_e$)]{The normalized region mutual information between the raw data ${\bf Y}_{\textrm{t}}$ and the approximation at level $L$. The mutual information monotonically decreases with increasing $L$ before being stagnant after reaching $L=6$. The rate of change becomes insignificant after $L=6$, suggesting the complete removal of separable faults from the approximation at level $6$. It can be observed that the trend is the same across all the excitation times ($T_e$)}
\label{fig:mutalinfo}
\end{center}
\end{figure}

We compute the RMI between the raw thermal image, ${\bf Y}_{\textrm{t}}$ and a subspace spanning the approximation at level $L$ for every pulse in the PRBS sequence. The resulting RMI is normalized by the mutual information of the raw image and then temporally averaged to obtain $M_{avg}(L)$ for a given basis function.
The resulting mutual information (Fig. \ref{fig:mutalinfo}) follows a monotonous trend as expected with reducing similarity index at each subsequent level. 
$M_{avg}(1)$ is the highest value obtained at the first level of decomposition and the difference between the successive level values remains constant until level $6$. 
However, from level $6$ onwards, this difference becomes relatively small and stays that way for subsequent levels. 
It can thus be deduced from this graph that level $6$ represents the maximum decomposition level for the test sample being analyzed. 
At this level, the most dominant fault has been completely detached from the background and the pictorial layers and hence the mutual information does not change significantly for higher levels of decomposition. 
This observation was ascertained by visual inspection of the approximation at level $6$, where the Hole $1$ was found to have been removed. 
%We represent the average mutual information for the $L^th$ level of decomposition as $MI_{avg}(L)$
%For the sake ease and simplicity only the results for the $db4$ are shown here.
%It is observed that the as we reach the $L=8$ the rate of decrease of mutual information is very small almost close to zero.
%It is to be considered that at every decomposition level in the DWT the approximation $A_L$ is broken down into $A_{L+1}$ and $D_{L+1}$.
%If the difference between the $MI_{avg}(L)$ and the $MI_{avg}(L+1)$ is small this means that removing the detail $D_{L+1}$ has not resulted in the removal of any significant information. 
%As the details and the approximations add up to form the raw acquired data so each of them contains a portion of the information contained in the raw data.
%If the Difference in the mutual information of the two adjacent approximations is very small this means that the detail removed contains very little portion of the information stored in the raw data and is thus unnecessary.
An other way to argue this is to use the scale(length in x and y direction) of the analyzing wavelet functions. This can be estimated in the x direction by $2^{L-1}N_x$ and in the y-direction by the $2^{L-1}N_y$ where $L$ is the level of wavelet decomposition.We can see that any value of L beyond $6$ would result in a very small length and can hence be thought of as noise and the $L=1 or 2$ result in big lengths and can be interpreted as capturing the background and illumination effects.As discussed in the Section \ref{sec:exp} the distance between the camera and the mural is fixed so the correspondence between the pixel in image and the real distance on the mural surface can be established via simple trigonometry. Hence a similar argument can be established to match different wavelet decomposition levels to size of faults and irregularities represented therein. 
The proposed level selection measure was tested for different basis functions excitation times ($T_e$) but similar trend (difference between successive levels) was observed albeit with minor variations in the parameter values as shown in the Figure \ref{fig:mutalinfo}  . 
The proposed measure presents the best compromise for the problem at hand and while it was developed based on laboratory generated defects, the obtained parameter (level of decomposition) can be utilized for real artworks having similar fault dimensions. 
At this point, it remains difficult to give a generic parameter, which caters for all types and dimensions of defect geometries. 
In a future work, it would be interesting to study the problem in the context of test sample characterization through its physical properties and to correlate with the findings of the proposed method. 

The framework discussed above allows an efficient selection of model parameters for the proposed algorithm. The useful subspace is then reconstructed by considering intermediate level details. The higher frequency details account for the noise, whereas the $L_{th}$ level approximation caters for the background affects. In the next section, the results obtained by application of the method to the test sample and the real artwork are presented along with elaborate discussion on the revealed fault maps.

%\begin{figure}
%\begin{center}
%\includegraphics[width=3in]{figures/rmi_tim_zubair.eps}
%\caption[Mutual Information of the wavelet basis bd4]{The normalized region mutual information between the raw data ${\bf Y}_{\textrm{t}}$ and the approximation at level $L$. The mutual information monotonically decreases with increasing $L$ before being stagnant after reaching $L=6$. The rate of change becomes insignificant after $L=6$, suggesting the complete removal of separable faults from the approximation at level $6$.}
%\label{fig:mutalinfo}
%\end{center}
%\end{figure}

\section{Results and Discussion}
\subsection{Application to Test Sample}

%%%%%%%
%%%%%%%

\begin{figure}
\centering
\subfigure[Raw thermal image ${\bf Y}_{\textrm{t}}$.]{
\includegraphics[width=3in]{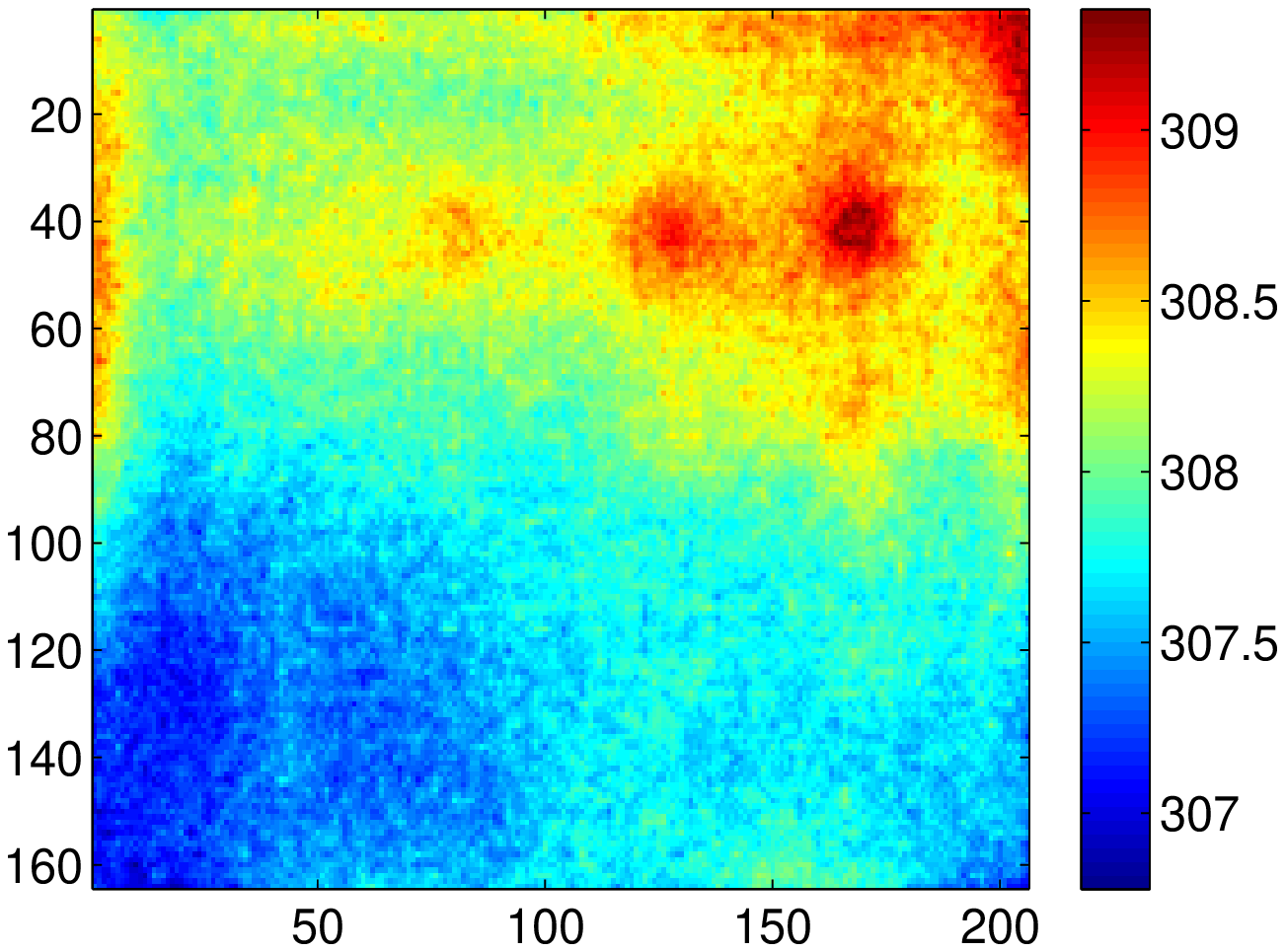}
\label{subfig:Initial}
}
\subfigure[$\bf{Y_{\textrm{det}}}$ revealing the hidden faults.]{
\includegraphics[width=2.5in]{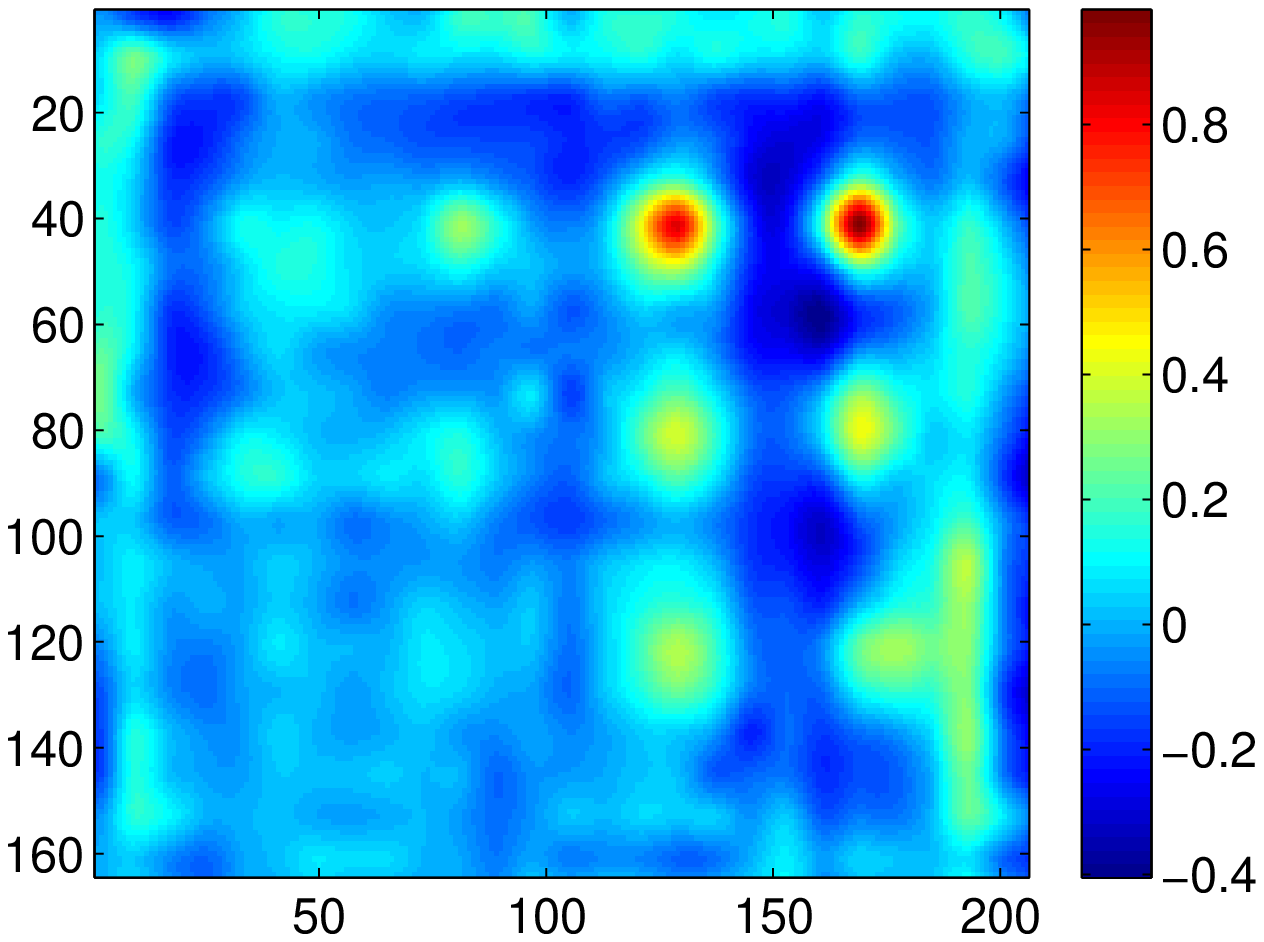}
\label{subfig:labdatares}
}
\subfigure[Skewness of the SVD based data reconstruction]{
\includegraphics[width=2.5in]{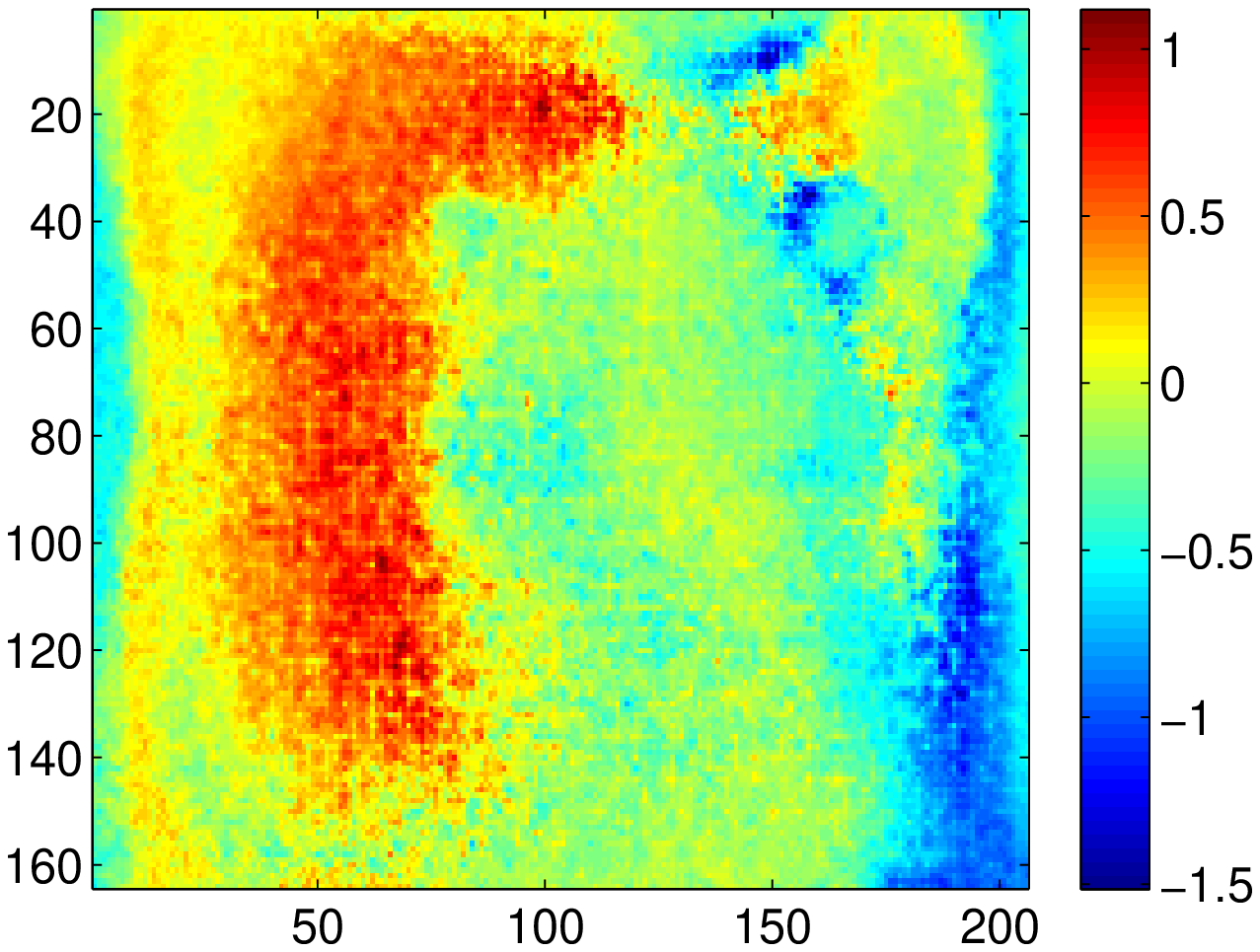}
\label{subfig:skewlabsvd}
}
\subfigure[Kurtosis of the SVD based data reconstruction]{
\includegraphics[width=2.5in]{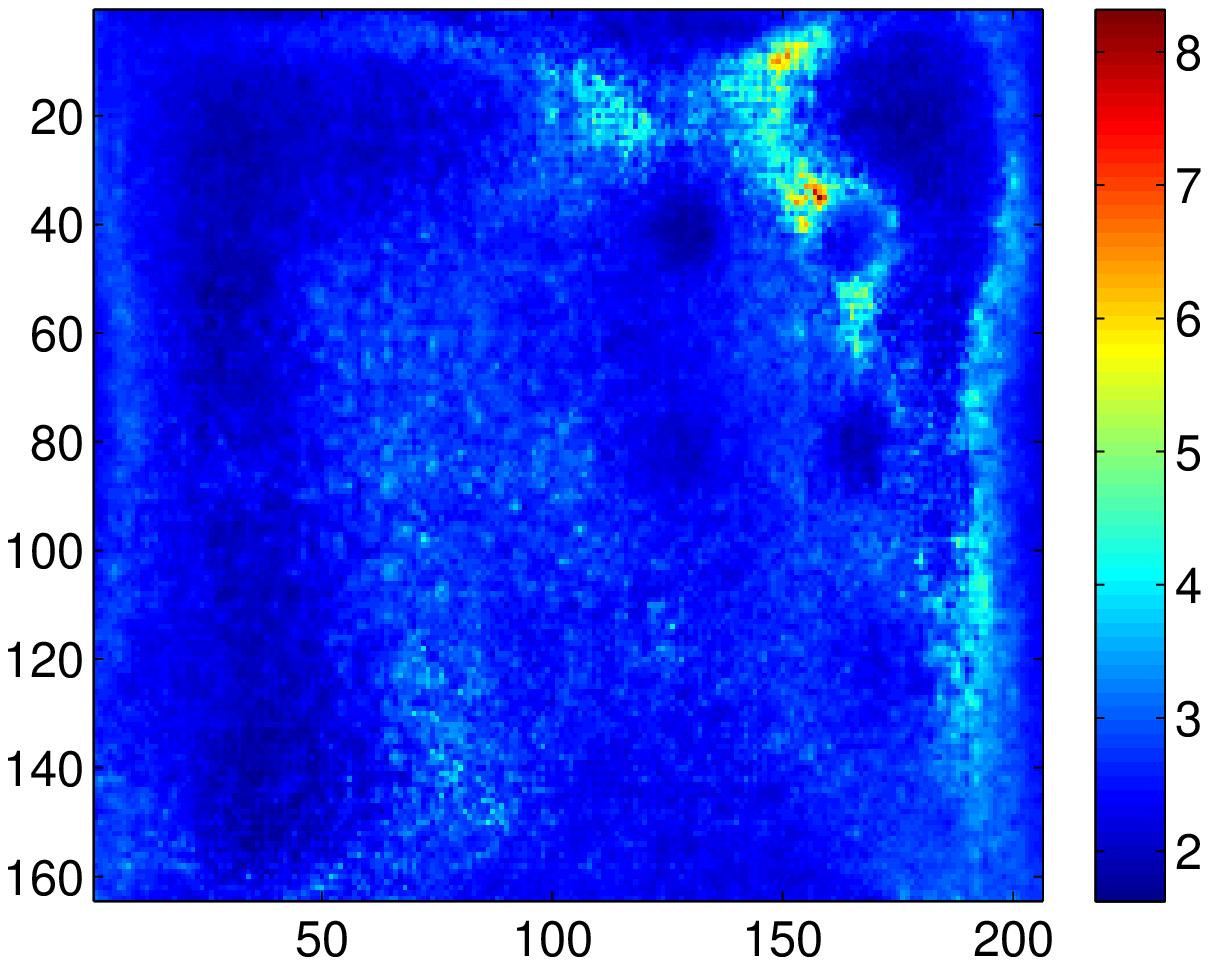}
\label{subfig:kurtlabsvd}
}
\subfigure[Fault map obtained by proposed algorithm]{
\includegraphics[width=2.5in]{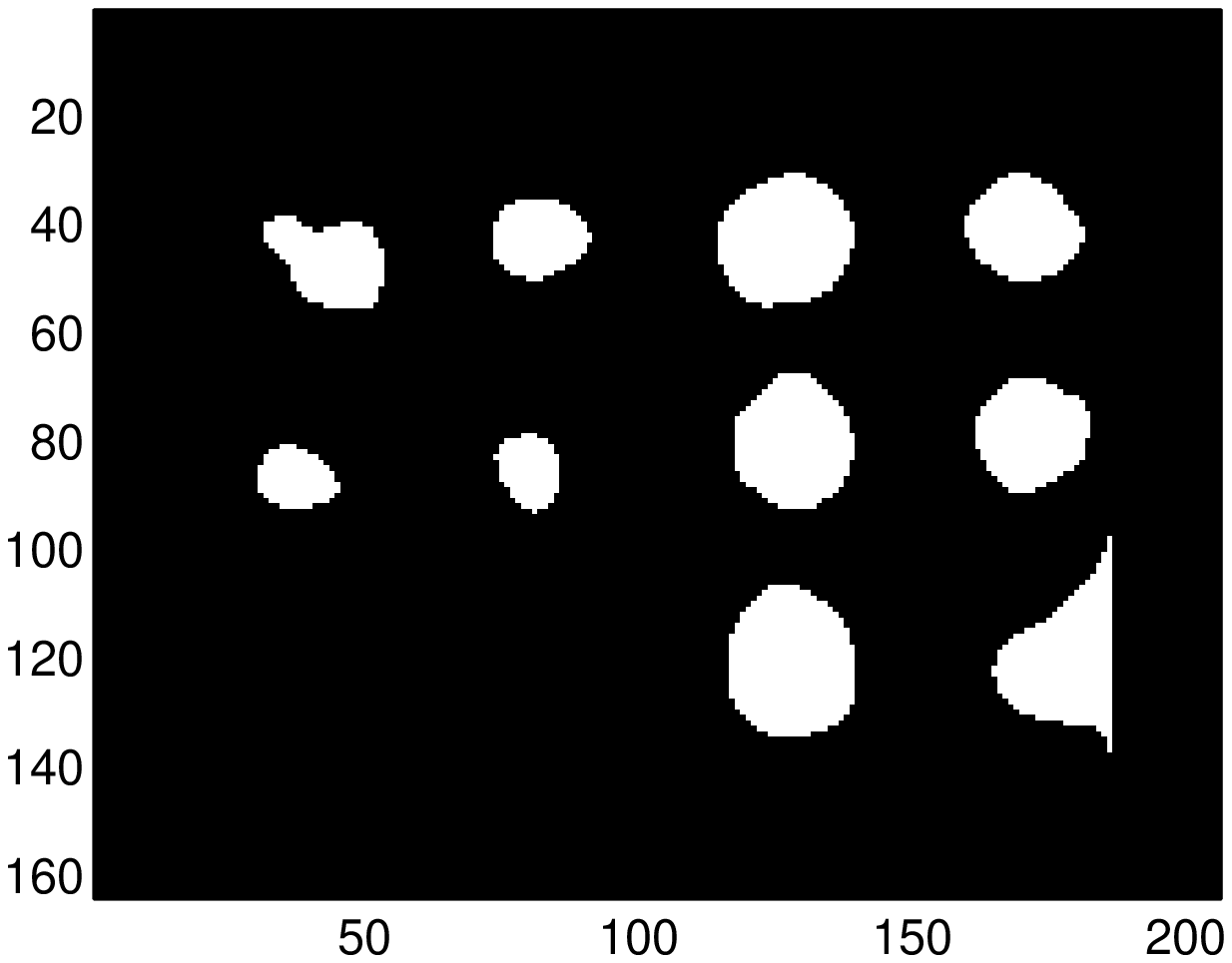}
\label{subfig:faultmap}
}
\caption[Result for test data developed in the laboratory]{The detection results for the laboratory generated test sample of Fig.\ref{fig:test_sample}): (a) raw data ${\bf Y}_{\textrm{t}}$ showing only Holes $1$ and $2$; (b) estimated $\bf{Y_{det}}$ with the proposed scheme, significantly improving identification of defects over the raw data; (c)-(d) the detection results of the SVD based approach showing the skewness and kurtosis, which do not reveal much information about the defects.(e) the fault map is calculated for the proposed algorithm by ignoring the edge artifacts and thresholding}
\label{fig:result_synth}
\end{figure}

%\subsection{Acquisition of Thermal Raw data}	
Fig. \ref{subfig:Initial} illustrates one of the raw images, captured on the test sample (Fig. \ref{fig:test_sample}), with the best visibility of smaller faults. 
It can be observed from this image that there is a strong influence of the illumination pattern on the acquired image. 
This pattern results from non-homogeneous illumination of the test sample by the two halogen lamps. 
An important processing task will thus be to suppress this pattern to reveal more relevant information regarding the defects. 
The most dominant hole, Hole $1$($4$mm depth, $10$ mm diameter), is easily observable on this raw image, while Hole $2$ also appears on the raw data albeit with reduced intensity. 
However, all the other holes have been masked by the strong illumination effect and the general texture of the test material. 
%\begin{center}
%\begin{flushleft}
%\begin{table}
%\caption{Cost function (Eq. \ref{eq:cstfn}) for a set of wavelet basis function, $\Phi$, tested to determine the optimal basis. The minimum cost function is obtained for $rbio6.8$.}
%\begin{tabular}{|c|c|c|c|c|c|c|c|c|}
%\hline
 %                  \multicolumn{9}{|c|}{Basis Functions}      \\ \hline
%\multirow{3}{*}{}  $\#$              & $1$ & $2$ & $3$ & $4$ & $5$ & $6$ & $7$ &$8$   \\ \cline{1-9} 
  %                 Basis  & $db4$  & $db10$  & $coif1$ & $coif4$ & $coif5$ & $sym4$ & $sym10$&$bior 2.2$  \\ \cline{1-9}
	%								Cost Function & $8.087$  & $7.951$  & $8.957$ & $8.239$ & $8.509$ & $7.761$ & $7.855$&$8.403$ \\ \hline
%\end{tabular}

%\begin{tabular}{|c|c|c|c|c|c|c|c|c|}
%\hline
 %                  \multicolumn{8}{|c|}{Basis Functions}      \\ \hline
%\multirow{3}{*}{}  $\#$  &$9$&$10$&$11$&$12$&$13$&$14$&$15$ \\ \cline{1-8} 
 %                  Basis  & $bior2.4$& $bior3.3$& $bior3.7$& $bior3.9$& $bior6.8$& $rbio3.7$& $rbio6.8$ \\ \cline{1-8}
	%								Cost Function & $8.009$& $8.903$& $8.7$& $8.683$& $7.766$& $9.1$& $7.685$\\ \hline
%\end{tabular}
%\label{tab:basisinfo}
%\end{table}
%\end{flushleft}
%\end{center}

\renewcommand{\tabcolsep}{1pt}
\begin{center}
%\begin{flushleft}
\begin{table}
\caption{Cost function values (Eq. \ref{eq:cstfn}) for a set of wavelet basis function for different excitation times ($T_e$), $\Phi$, tested to determine the optimal basis.The optimal values for different excitation times are in boldface.}
\begin{tabular}{|c|c|c|c|c|c|c|c|c|c|c|c|c|c|c|c|c|c|}
\hline
                  
\multirow{6}{*}{}   
                   Basis  & $db4$  & $db10$  & $coif1$ & $coif4$ & $coif5$ & $sym4$ & $sym10$&$bior 2.2$ & $bior2.4$& $bior3.3$& $bior3.7$& $bior3.9$& $bior6.8$& $rbio3.7$& $rbio6.8$ \\ \cline{1-16}
									$T_e=0.5s$ & $9.837$ & $9.583$ & $\mathbf{8.340}$ & $9.346$ & $9.563$ & $8.894$ & $9.088$ & $8.658$&$8.5910$&$11.0385$&$10.0823$&$9.8274$&$8.9697$&$8.9650$&$8.9288$ \\ \hline
									$T_e=1s$&$8.9912$&$8.9888$&	$7.6042$&$	8.383$&$	8.7456$&$	\mathbf{7.5807}$&$	8.1286$&$	7.9584$&$7.7010$&$	10.8070$&$	10.0097$&$	9.7387$&$	7.9985$&$	8.9699$&$	7.9613$ \\ \hline
									$T_e=2s$&$7.8232$&$	7.3634$&$	6.8023$&$	7.8996$&$	7.7596$&$	\mathbf{6.097}$&$	7.2485$&$	6.922$&$6.7302$&$	8.4880$&$	8.1136$&$	8.1599$&$	7.1465$&$	8.5845$&$	7.1146$ \\ \hline
									$T_e=5s$ & $8.087$  & $7.951$  & $8.957$ & $8.239$ & $8.509$ & $7.761$ & $7.855$&$8.403$& $8.009$& $8.903$& $8.7$& $8.683$& $7.766$& $9.1$& $\mathbf{7.685}$ \\ \hline
									$T_e=10s$&$8.9017$&$	8.9301$&$	\mathbf{7.8784}$&$	8.4588$&$	8.6935$&$	7.9610$&$	8.2871$&$	7.9732$&$7.9401$&$	10.5622$&$	9.6351$&$	9.3328$&$	8.1533$&$	8.8919$&$	8.1085$ \\ \hline
\end{tabular}

%\begin{tabular}{|c|c|c|c|c|c|c|c|c|}
%\hline
                   
%\multirow{6}{*}{}  
 %                  Basis  & $bior2.4$& $bior3.3$& $bior3.7$& $bior3.9$& $bior6.8$& $rbio3.7$& $rbio6.8$ \\ \cline{1-8}
	%								\tcr{$T_e=0.5s$}&$8.5910$&$11.0385$&$10.0823$&$9.8274$&$8.9697$&$8.9650$&$8.9288$ \\ \hline
		%							\tcr{$T_e=1s$}&$7.7010$&$	10.8070$&$	10.0097$&$	9.7387$&$	7.9985$&$	8.9699$&$	7.9613$ \\ \hline
			%						\tcr{$T_e=2s$}&$6.7302$&$	8.4880$&$	8.1136$&$	8.1599$&$	7.1465$&$	8.5845$&$	7.1146$\\ \hline
				%					$T_e=5s $& $8.009$& $8.903$& $8.7$& $8.683$& $7.766$& $9.1$& $7.685$\\ \hline
					%				\tcr{$T_e=10s$}&$7.9401$&$	10.5622$&$	9.6351$&$	9.3328$&$	8.1533$&$	8.8919$&$	8.1085$ \\ \hline
%\end{tabular}
\label{tab:basisinfo}
\end{table}
%\end{flushleft}
\end{center}

In the first instance, we consider application of the proposed method to the laboratory developed test sample (Fig. ~\ref{fig:rawdatacube}). 
The first parameter that needs to be selected is the set of basis functions for data decomposition and reconstruction. 
In this study, we tested different standard wavelet basis, a subset of which is listed in Table \ref{tab:basisinfo}. 
Adopting the criterion developed in Eq. \ref{eq:cstfn}, the proposed measure for wavelet basis selection is also presented in Table \ref{tab:basisinfo}. 
%In the current study we used the Laboratory developed data set depicted in the figure \ref{fig:rawdatacube} to choose the optimal basis function.The results obtained are shown in the figure \ref{fig:measure}.
Although the analysis was performed for different excitation times, results and the fault maps for only one ($T_e=5s$) are displayed (Figure \ref{fig:result_synth}) due to the shortage of space. The following discussion hence discusses this one case.
The cost function has comparable values for a number of different basis functions and the minimum value is achieved for the basis function $15$, as shown in Table \ref{tab:basisinfo}, i.e., the reverse biorthogonal rbio $6.8$. 
It may be argued that the selected basis is not uniquely the optimal basis, which will provide the best results in all the scenarios, but it has been derived systematically rather than through trial and error. 
All the subsequent analysis on both the laboratory developed sample and the real art works will be carried out using this optimal basis set. 
% Hence, this basis function is selected as the best basis function and is used throughout the paper to analyze both the Laboratory developed test data and the Mural 1 and 2.

The application of the proposed algorithm to the test sample of Fig. \ref{fig:test_sample} leads to the detection results illustrated in Fig.~\ref{subfig:labdatares}. The data are decomposed till level $6$ and the useful subspace is reconstructed from the details obtained at level $3$ to level $6$. 
%yields the results as shown in Figure ~\ref{subfig:labdatares}. 
In order to highlight the efficacy of the proposed algorithm, a comparison with the previously developed SVD based approach \cite{vrabie2012active} is performed. 
The SVD based subspace decomposition approach works on temporal signatures for each pixel and reconstructs the useful subspace after subtraction of a reference subspace, corresponding to the first singular value, and the high frequency noise subspace. 
The higher order statistics of skewness and kurtosis are then computed over the useful subspace temporally to obtain the final detection images demonstrated in Fig. \ref{subfig:skewlabsvd}-\ref{subfig:kurtlabsvd}. 
The results of the proposed approach clearly outperform the results of the SVD based approach where none of the defects, except a few traces of Hole $1$, appear in the detection results.
A comparison with the raw data (Fig.~\ref{subfig:Initial}) suggests that the SVD based approach was unsuccessful in achieving a good separation between the defects and the background. 
This can primarily be attributed to the inherent nature of SVD which is based on diagonalization using second order statistics. 
Moreover, the SVD based approach involves temporal processing of the data while considering spatial independence. 
The temporal evolutions of faulty and non-faulty pixels in the acquired image follow a very similar trend and while, there are subtle variations in their respective temporal behaviours, the difference is not significant to establish a pronounced distinguishability. 
The suppression of the subspace corresponding to the first singular value therefore suppress useful information related to the defects. 
%The excitation time,$T_e$, for this experiment is set to be $5s$. Similarly to draw a comparison between the two decomposition methods, namely the wavelet decomposition and the SVD the results obtained by the combination of SVD and HOS,as proposed in \cite{vrabie2012active} are also included.  

A comparison of the results of the proposed approach (Fig.~\ref{subfig:labdatares}) with the raw dataset (Fig.~\ref{subfig:Initial}) clearly shows that the proposed approach allows identification of most of the defects in the test material with a better spatial localization. 
In particular, of the total $12$ holes, Holes $1$-$10$ are visible now after processing through the proposed system. 
The faults corresponding to Holes $1$ and $2$ are the most dominant followed by Holes $3$, $5$, $6$, $9$ and $10$.
The Holes $4$, $7$ and $8$ appear with relatively lesser intensities as compared to other faults due to their dimensions. 
Finally, the weakest faults Holes $11$ and $12$ are not detected by our proposed approach. 
It is pertinent to recall that the thermal radiography is a technique, which works well closer to the surface and these weaker holes are probably too small and too far from the surface. 
Their non-detectability is thus more to do with the physical limitation of the acquisition process than the proposed processing scheme. Moreover, for a given depth, the intensity of the detected faults enhances with increasing diameters and similarly for a given diameter, the intensity reduces as the faults move farther from the material surface. 
Although, at this stage, it is not the interest to accurately quantify the defects in terms of their dimensions but the relative intensities of different holes in the final detection parameter is a potential indicator of the fault dimension (depth and diameter). This opens up the possibility of exploring indirect fault quantification oriented approach in the future.

In order to give a qualitative measure for the utility of the proposed approach, we compare the Signal to Noise Ratio (SNR) for different defects in the raw data against the SNR in the processed detection results. 
In order to compute the SNR after the application of the proposed algorithm, the signal power for a given hole is computed in a window centered around that hole, whereas the noise power is computed by averaging the powers in different background windows not containing the faults, being the same with $\bf{B}_{avg}$ used in Eq.~\ref{eq:cstfn}. This process can be represented by the equation \eqref{SNRwd} .

The value $jopt$ is the optimal basis for the wavelet decomposition as described in the section \ref{subsec:Paramopt} and $k$ is the hole number thus $k=1,2,\cdots ,12$. The SNR for the raw data ($SNR_{raw}$) is calculated in a similar way by replacing the $\bf{Y_{det}(win)_{(k,j-opt)}}$ with $\mathcal{Y}(win)_{k}$ as represented in \eqref{SNRraw} . To formulate a measure of the performance of the proposed algorithm denoted $SNR_{imprv}$ we calculate the improvement in the SNR resulting from the application of the algorithm using the equation \eqref{SNRimp}.

\begin{subequations} 
\begin{align}
SNR_{wd}=\frac{\bf{Y_{det}(win)}_{(k,j-opt)}}{\bf{B}_{avg,jopt}}  \label{SNRwd} \\
SNR_{raw}=\frac{\bf{\mathcal{Y}(win)}_{(k)}}{\bf{B}_{avg,jopt}}   \label{SNRraw} \\
SNR_{imprv}=SNR_{wd}-SNR_{raw}    \label{SNRimp}
\end{align}
\label{eq:SNR}
\end{subequations}

These $SNR_{imprv}$ values for various excitation times are shown in Table \ref{tab:holeinfo}. 

It can be observed that the SNRs for most of faults were very low in the raw data and the Hole $1$, representing the most dominant hole, was almost at the same level as the background. 
The other holes scored progressively lower with respect to SNR in the raw data. 
A marked improvement is obtained after processing the data through the proposed algorithm as seen in the last column of Table \ref{tab:holeinfo}. 
Hole $1$ now has an SNR of $9.7$ dB whereas most of the other holes which were not visible in the raw data have been significantly enhanced to higher level SNRs. 
This clearly demonstrates the strength of the proposed approach to extract the useful information from the raw thermal images. 
Some of the lower holes, however, demonstrate lower SNRs but still show some visible traces in the detected image.
The improvement in the SNR achieved by the application of the proposed methodology on the rest of the excitation times ($T_e$) has been included in the Table \ref{tab:holeinfo}. It demonstrates that the results obtained are quite similar to the case discussed above and that the proposed algorithm works effectively.
%Zubair include two columns in Table 1. and add text above in line with the table results on SNR.

%The reason for this being the depths and the diameters,Figure ~\ref{test_sample_b}. The exact relationship mapping the depths and the diameters to the detection strength was not the scope of this study. The objective was to achieve enhanced detection of faults as compared to the previously used method namely the combination of  SVD and HOS.
%The figures \ref{subfig:skewlabsvd} and  \ref{subfig:kurtlabsvd} show that the combination of SVD and HOS fails to detect any holes either in the skewness or the kurtosis of the SVD based reconstruction. The reasons for this are the inherent weakness in the SVD as explained previously in section \ref{subsec:SVDweakness}.

%\begin{figure}
%\begin{center}
%\includegraphics[width=2.5in]{figures/measure3}
%\caption{The values of the Equ;\ref{eq:cstfn} for different basis functions. The X-axis values ranging from 1 to 15 represent different basis functions. Basis function number 15 gives the minimum most value and hence is nominated the best basis function.}
%\label{fig:measure}
%\end{center}
%\end{figure}  
% Please remember to add \use{multirow} to your document preamble in order to suppor multirow cells
% Please remember to add \use{multirow} to your document preamble in order to suppor multirow cells
 
\subsection{Application on the real artwork}
\label{subsec:realdatastateofart}
\begin{figure}
\begin{center}
	\subfigure[Raw thermal image ${\bf Y}_{\textrm{t}}$]{
	\includegraphics[width=7cm]{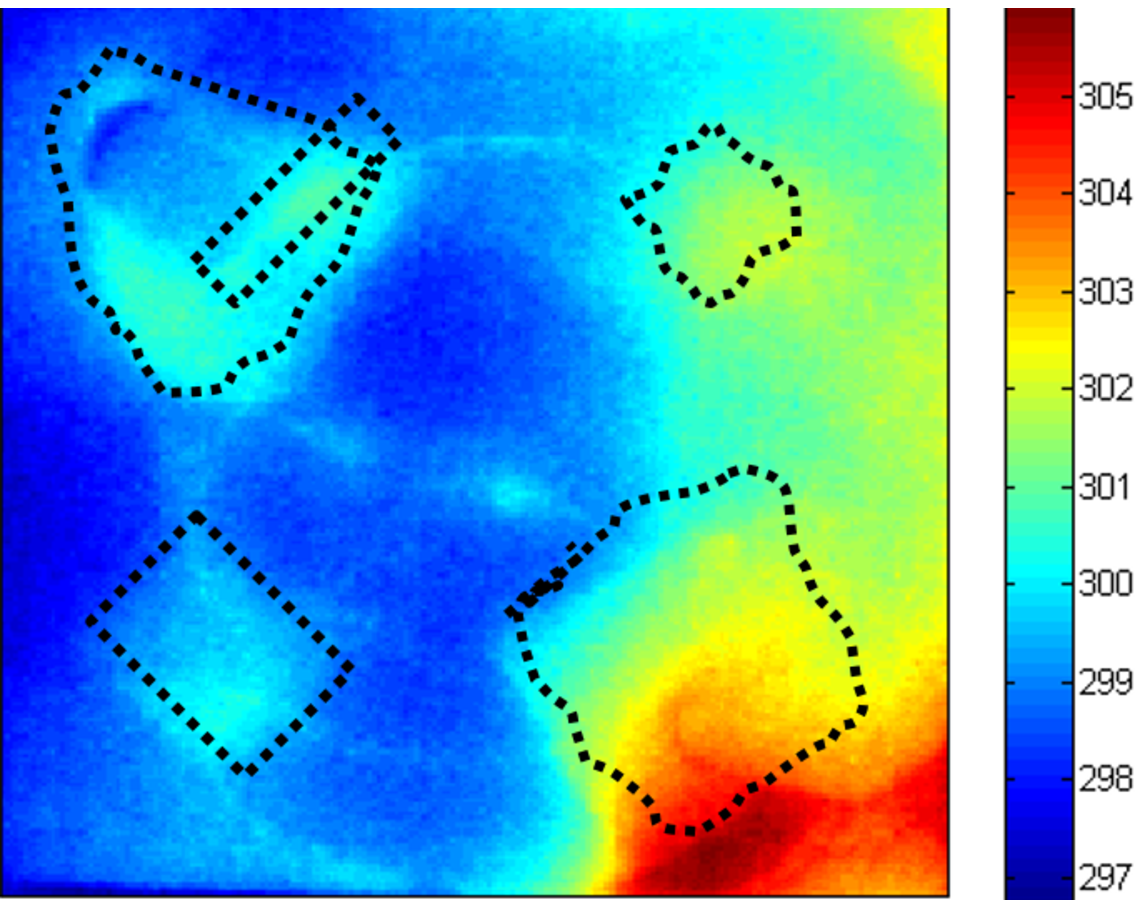}
	\label{subfig:stcris_orig}
	}
	\subfigure[Detection results $\bf{Y_{\textrm{det}}}$ of the proposed scheme]{
	\includegraphics[width=7cm]{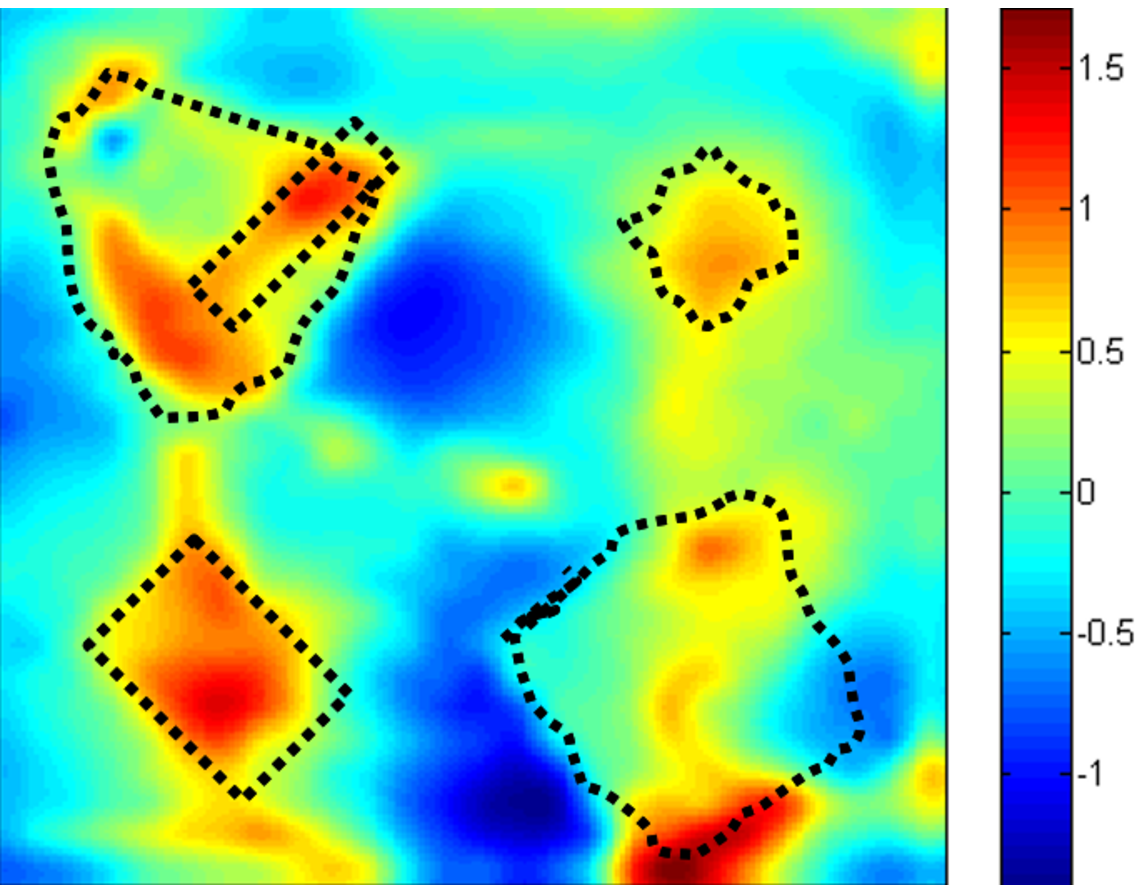}
	\label{subfig:stcris_rslt}
	}
	\subfigure[Skewness of the SVD based data reconstruction]{
	\includegraphics[width=7cm]{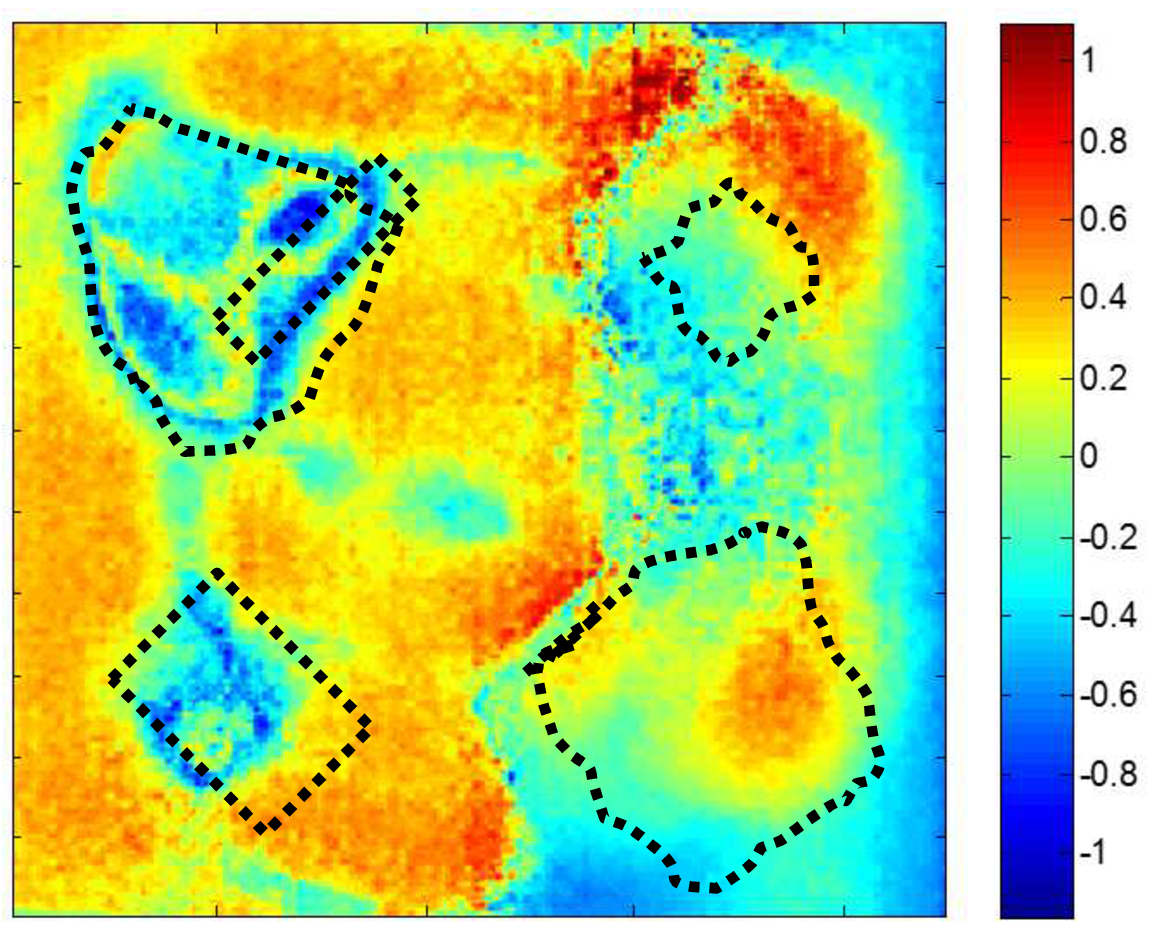}
	\label{subfig:Skew}
	}
	\subfigure[Kurtosis of the SVD based data reconstruction]{
	\includegraphics[width=7cm]{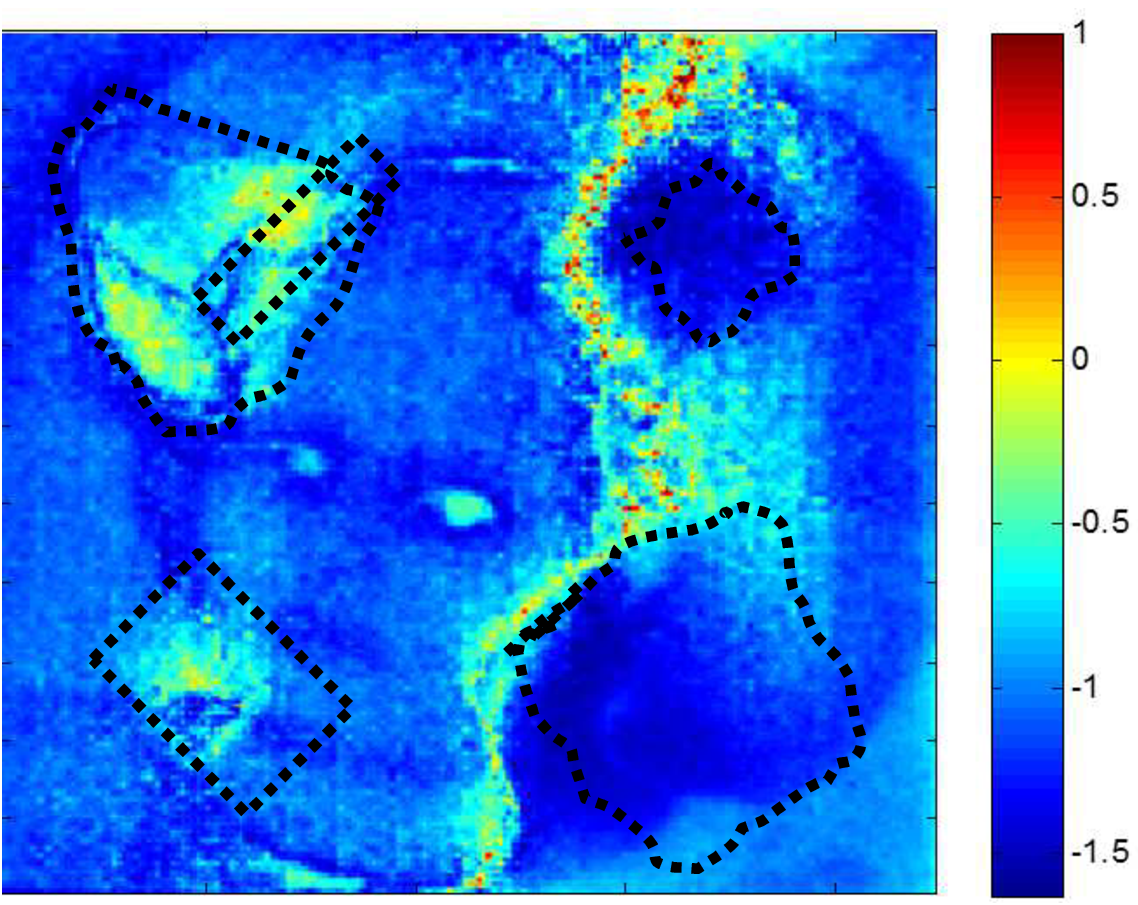}
	\label{subfig:Kurt}
	}
	\caption{Detection parameter for the Mural $1$: (a) Raw thermal image at an acquisition instant providing best visibility of faults in raw data; (b) Detection parameter of proposed scheme improving on the raw results; (c)-(d) Results of SVD based approach showing skewness and kurtosis. The contours mark the edges of the faults $\bf{A}$ to $\bf{E}$ (Fig. \ref{fig:test_sample_stchris}). $4$ defects are revealed by the proposed approach in (b) with some anomalies corresponding to the eye region of Mural $1$.}
	\end{center}
	\end{figure}

After validation of the proposed method on the laboratory generated test sample, we focus our attention to the real artwork. All the parameters of the proposed algorithm including the wavelet basis and the decomposition levels for Mural $1$ are kept same as those for the laboratory generated test sample. Fig. \ref{subfig:stcris_orig} shows the best raw image available for Mural $1$. 
The detection results obtained for the Mural $1$ are shown in Fig. \ref{subfig:stcris_rslt} which demonstrate a very good detection of all faults. 
Faults $C$ and $D$ are highlighted with same intensities. 
This is very interesting since they are both at the same, $3$ mm, depth. 
Fault $A$ is highlighted in the bottom with a higher intensity. It is interesting to note that this fault manifests variable detection intensities in its proximity, a fact attributable to the varying depth of this defect from $3$ mm to $10$ mm. 
Fault $B$, which is deeper at 5 mm, shows a lighter intensity which correspond to the center of the fault $A$. 
A region around the eyes in Mural $1$ is also detected in the result, which is a false alarm in this case. 
The defect $E$ is not detectable by the proposed approach, owing partially to its deep location ($10$ mm) from the surface of the material. 
However, a slight heterogeneity on the upper right part of the fault $D$ might be attributable to this fault.
Compared with the state-of-the art method (results shown in Figs. \ref{subfig:Skew} and \ref{subfig:Kurt}), the shapes of the faults are better identified, especially for the ones closest to the front side ($C$, $D$ and $B$). 
Moreover, this systematic identification of faults is available through a single output of the proposed method. This is in contrast to the skewness, which identifies only the faults $C$ and $D$ and to the kurtosis, which identifies only the faults $A$ and $B$, albeit with a poor shape.

The proposed method was tested on other data sets as well with success. This non-invasive method, thus allows us to characterize the faults in artworks using a simple setup with a heating distributed over the time by a PRBS excitation and a strong processing strategy. The approach can be adapted to other types of materials with minimal modifications. 

\begin{figure}
\begin{center}
\subfigure[The raw data for Mural $2$]{\includegraphics[width=2.2in]{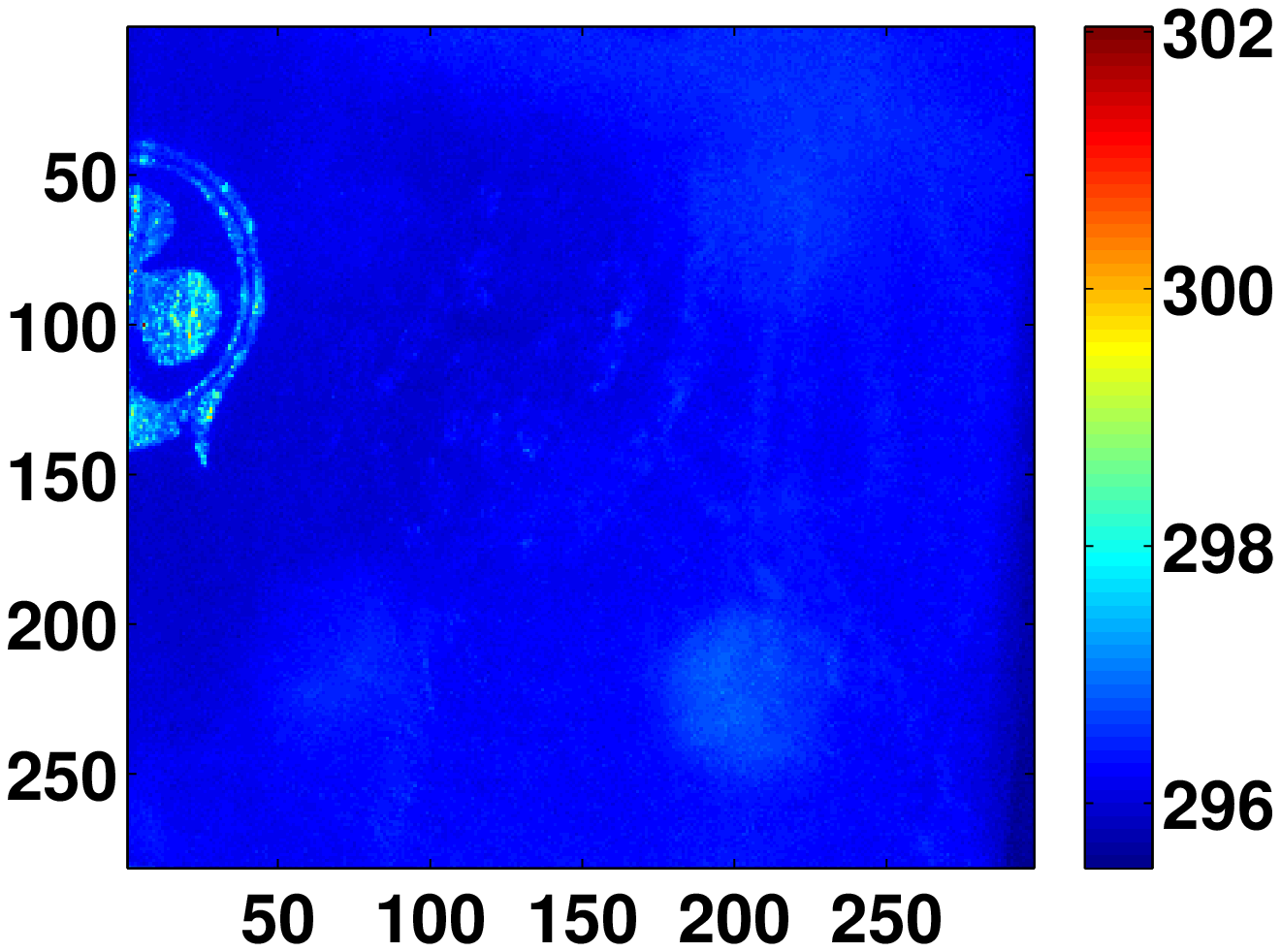}}
\label{subfig:stavinorig}
\subfigure[The fault map for Mural $2$.]{\includegraphics[width=2in]{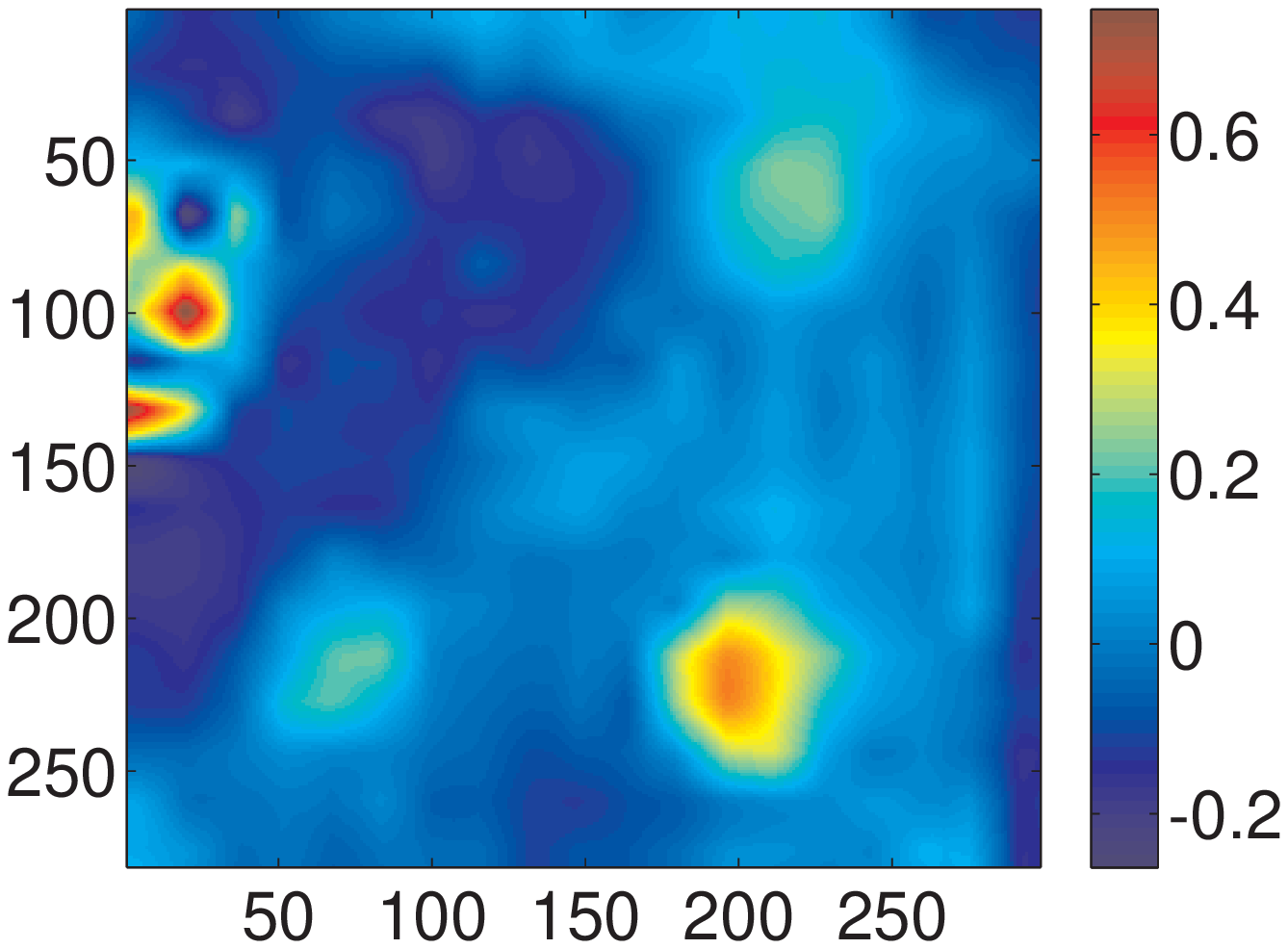}}
\label{subfig:stavinrslt}
\subfigure[The corrected detection after removing correlated data.]{\includegraphics[width=2in]{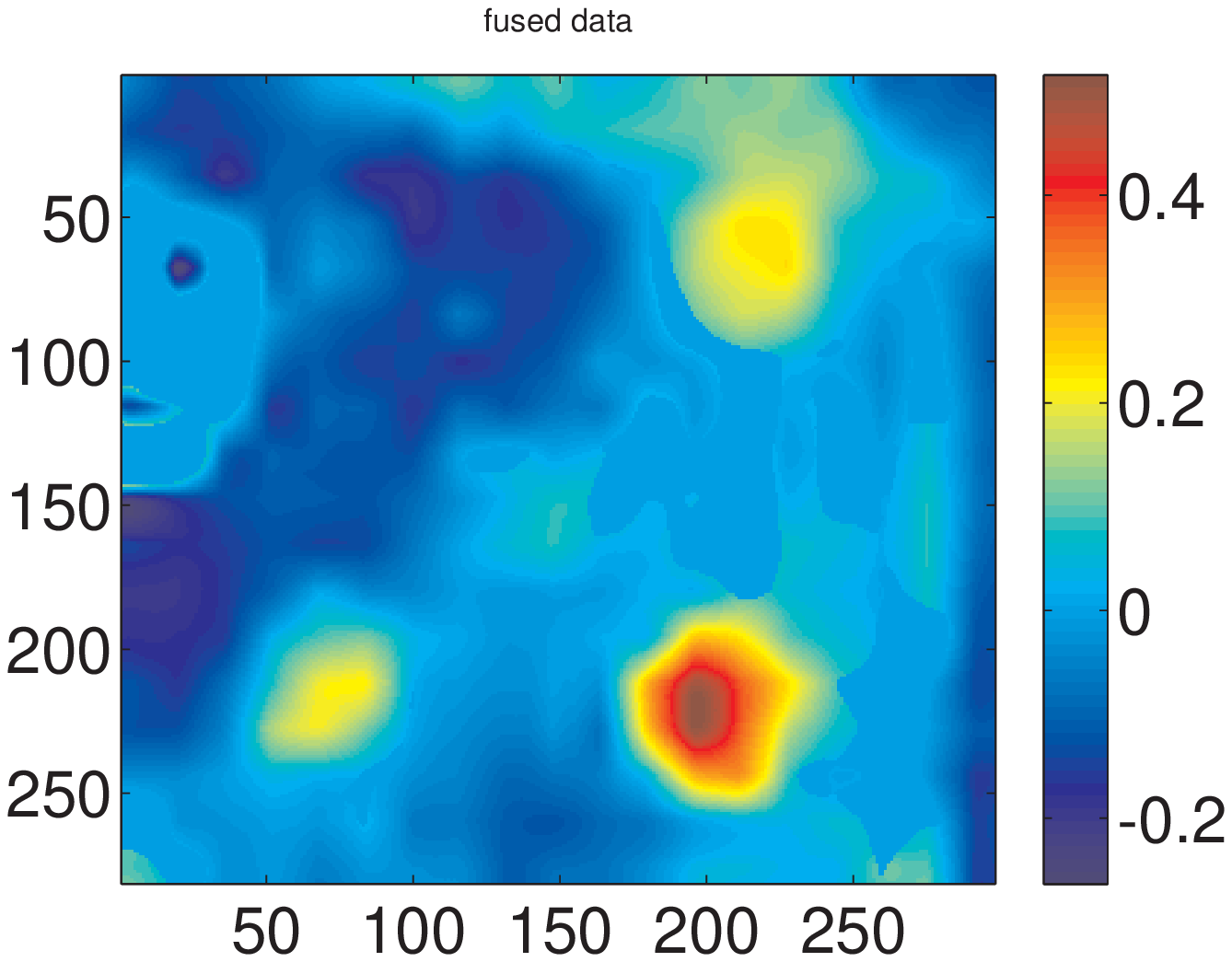}
\label{subfig:stavinrsltcr}}
\caption[Results of the application of proposed algorithm on Mural 2]{The detection results for Mural $2$:(a)The raw data for the $Mural 2$ , gold foil is visible (b) the detection parameter of the proposed algorithm depicting four anomalies; (c) the correlation corrected image allowing removal of the anomaly on top left of (a) that was associated with a gold foil.}
\end{center}
\end{figure}

\begin{figure}
\begin{center}   
		\subfigure[Comparison of the Timeseries of Fault, Background and the Gold Foil obtained after proposed algorithm]{%
\includegraphics[width=7cm]{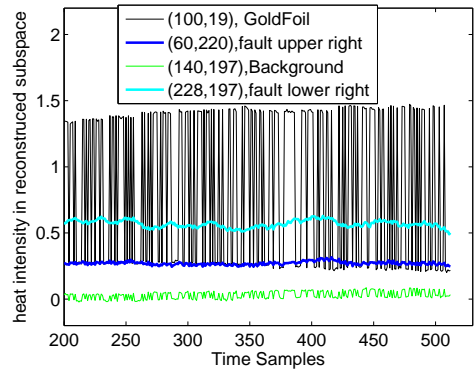}
			\label{subfig:prbsseq}
    }
    
    \subfigure[PRBS excitation sequence used for the Mural 2 Analysis]{%
      \includegraphics[width=7cm]{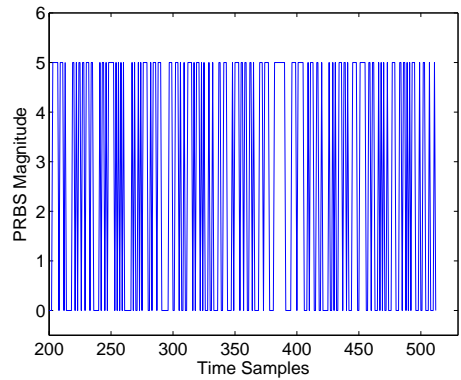}
			\label{subfig:corelation}
    }
			
\caption{Temporal evolution of the reconstructed subspace at four different spatial locations for the Mural $2$ including three anomalies (top left, top right, top left) and a background location. The time series corresponding to the anomaly on top left exhibits a high correlation with the excitation sequence and this anomaly thus represents a highly reflective surface.}
\label{subfig:prbs}
	\end{center}
  \end{figure}

The proposed algorithm is next applied to Mural $2$ to obtain the detection results as shown in Fig.~\ref{subfig:stavinrslt}. While, the raw image depicted a strong anomaly in the top left corner (Fig. \ref{subfig:stavinorig}), the detection results show three additional anomalies. In order to obtain a closer inspection of the identified anomalies, the temporal evolution of the the reconstructed subspace is plotted in Fig.~\ref{subfig:prbs} for four different regions including two principal anomalies on the right side (top and bottom), a background location and the anomaly on the top left. It was observed that the anomaly on the top left exhibits a very high temporal correlation with the excitation sequence, whereas the temporal correlations for the other anomalies are relatively lower. The results are then post adjusted by setting a correlation based threshold on the detection results to remove the anomalies that are highly correlated with the excitation sequence. The results after applying this correction are demonstrated in Fig.~\ref{subfig:stavinrsltcr}. The three anomalies originally identified are retained in the detection parameter whereas the fourth anomaly has been subdued to a large extent. This fourth anomaly was actually due to a gold foil in that spatial vicinity which resulted in a high reflection and thus a higher correlation with the excitation sequence. The analysis thus allows us not only to detect the defects in the real artworks with a better separation from the background but also to identify the anomalies that may result from highly reflective surfaces. 
Out of these regions the one on the top left is an anomaly caused by a gold foil coating on the surface of the specimen being analysed.
On closer inspection we find that the first point is not a conventional fault but is a piece of gold foil on the pictorial layer. This leads to its distinct behavior when we plot the time series specified by this point in $Y_{recon}$ as shown in Figure ~\ref{subfig:prbs}. The point visually shows marked similarity between the PRBS excitation sequence and the time-series $Y_{recon}(19,100)$. When we take the correlation of all the four time-series with the PRBS excitation sequence the result is depicted in Figure ~\ref{subfig:corelation}. Two points which yield a very high correlation are the $Y_{recon}(19,100)$ and the $Y_{recon}(197,140)$ the former being the gold foil and later the background.Based on the marked difference in the correlation of the different time-series we apply thersh-holding on the maximum value of the correlation coefficient to eliminate the effect of reflective materials (gold foil)and thus obtain an improved estimate of the fault location as shown in Figure ~\ref{subfig:stavinrsltcr}.   

%\begin{table}
%\caption{\tcb{Dr. Amir, Sir I think this tabel is also redundant based on the modified figure 11}}
%\begin{tabular}{|c|c|c|c|c|}
%\hline
%                   \multicolumn{5}{|c|}{St. Stavin fault and background Locations used for analysis depicted in Figure \ref{fig:stavin}}      \\ \hline
%\multirow{2}{*}{}  $x$ & $19$ & $197$ & $197$ & $220$ \\ \cline{1-5} 
 %                  $y$ & $100$  & $140$  & $228$ & $60$  \\ \hline
%\end{tabular}
%\label{tab:ststavininfo}
%\end{table}
 \section{Conclusion}
Non-destructive testing of materials is a highly developed field where different techniques are employed for automated characterization of test materials. One particular type of material that is highly sensitive to rupture is ancient artwork. Photo-radiometry can be employed to test these artworks, where the material is heated and the resulting thermal response is captured by a thermal infrared camera. In this paper, a pseudo-radom binary sequence based excitation, as opposed to the classical pulse excitation, is employed which prevents excessive heating of the material thus reducing the risk of irreversible damage to the artworks. Using simplified thermal models, the faults in the material should induce an irregularity in the acquired thermal image at their spatial locations. However, owing to different factors such as the background pictorial layer, illumination conditions, the measurement noise and the diffusion, the faults are not readily identifiable on the raw acquired images. While, the faults of larger dimensions are visible in the raw data, they are at similar intensity levels as the background. The weaker faults are masked by the afore mentioned factors and are not visible at all in the raw data. Considering the spatial non-stationarity in the acquired raw data, we developed an algorithm based on the wavelet subspace decomposition. Appropriate selection criteria for selection of different parameters like the basis selection and the decomposition level selection were also proposed in this paper. The application on a laboratory generated test sample allowed establishing these parameters and the final subspace obtained represented the fault maps in the spatial domain, thereby allowing characterization of different faults. A significant enhancement in the SNR of the faults was obtained in the detected subspace as compared to the raw data. The method was also demonstrated to successfully identify faults in real artworks. It was also observed that the intensities of the detected fault maps are proportional to the fault dimensions (diameter and depth), a potential indicator for quantification of faults in future work. In brief, this work integrates a simple non-invasive material testing setup with a post-processing strategy, enabling an efficient spatial localization of faults in the material.

\bibliographystyle{IEEEtran}
\bibliography{zub}

\begin{thebibliography}{10}
\providecommand{\url}[1]{#1}
\csname url@samestyle\endcsname
\providecommand{\newblock}{\relax}
\providecommand{\bibinfo}[2]{#2}
\providecommand{\BIBentrySTDinterwordspacing}{\spaceskip=0pt\relax}
\providecommand{\BIBentryALTinterwordstretchfactor}{4}
\providecommand{\BIBentryALTinterwordspacing}{\spaceskip=\fontdimen2\font plus
\BIBentryALTinterwordstretchfactor\fontdimen3\font minus
  \fontdimen4\font\relax}
\providecommand{\BIBforeignlanguage}[2]{{%
\expandafter\ifx\csname l@#1\endcsname\relax
\typeout{** WARNING: IEEEtran.bst: No hyphenation pattern has been}%
\typeout{** loaded for the language `#1'. Using the pattern for}%
\typeout{** the default language instead.}%
\else
\language=\csname l@#1\endcsname
\fi
#2}}
\providecommand{\BIBdecl}{\relax}
\BIBdecl

\bibitem{Baker-Jarvis1994}
J.~Baker-Jarvis, M.~D. Janezic, P.~D. Domich, and R.~G. Geyer, ``Analysis of an
  open-ended coaxial probe with lift-off for nondestructive testing,''
  \emph{Instrumentation and Measurement, IEEE Transactions on}, vol.~43, no.~5,
  pp. 711--718, 1994.

\bibitem{Vasic2004}
D.~Vasic, V.~Bilas, and D.~Ambrus, ``Pulsed eddy-current nondestructive testing
  of ferromagnetic tubes,'' \emph{Instrumentation and Measurement, IEEE
  Transactions on}, vol.~53, no.~4, pp. 1289--1294, 2004.

\bibitem{Bernieri2000}
A.~Bernieri, G.~Betta, G.~Rubinacci, and F.~Villone, ``A measurement system
  based on magnetic sensors for nondestructive testing,'' \emph{Instrumentation
  and Measurement, IEEE Transactions on}, vol.~49, no.~2, pp. 455--459, 2000.

\bibitem{Benedetti2006}
M.~Benedetti, M.~Donelli, A.~Martini, M.~Pastorino, A.~Rosani, and A.~Massa,
  ``An innovative microwave-imaging technique for nondestructive evaluation:
  applications to civil structures monitoring and biological bodies
  inspection,'' \emph{Instrumentation and Measurement, IEEE Transactions on},
  vol.~55, no.~6, pp. 1878--1884, 2006.

\bibitem{Karagiannis2011}
G.~Karagiannis, D.~S. Alexiadis, A.~Damtsios, G.~D. Sergiadis, and
  C.~Salpistis, ``Three-dimensional nondestructive "sampling" of art objects
  using acoustic microscopy and time--frequency analysis,''
  \emph{Instrumentation and Measurement, IEEE Transactions on}, vol.~60, no.~9,
  pp. 3082--3109, 2011.

\bibitem{Faifer2011}
M.~Faifer, R.~Ottoboni, S.~Toscani, and L.~Ferrara, ``Nondestructive testing of
  steel-fiber-reinforced concrete using a magnetic approach,''
  \emph{Instrumentation and Measurement, IEEE Transactions on}, vol.~60, no.~5,
  pp. 1709--1717, 2011.

\bibitem{Ricci2012}
M.~Ricci, L.~Senni, and P.~Burrascano, ``Exploiting pseudorandom sequences to
  enhance noise immunity for air-coupled ultrasonic nondestructive testing,''
  \emph{Instrumentation and Measurement, IEEE Transactions on}, vol.~61,
  no.~11, pp. 2905--2915, 2012.

\bibitem{Wilson2012}
K.~Wilson, K.~Homan, and S.~Emelianov, ``Biomedical photoacoustics beyond
  thermal expansion using triggered nanodroplet vaporization for
  contrast-enhanced imaging,'' \emph{Nature communications}, vol.~3, p. 618,
  2012.

\bibitem{Bagava2013}
S.~Bagavathiappan, B.~Lahiri, T.~Saravanan, J.~Philip, and T.~Jayakumar,
  ``Infrared thermography for condition monitoring – a review,''
  \emph{Infrared Physics \& Technology}, vol.~60, pp. 35 –-- 55, 2013.

\bibitem{Qingju2015}
T.~Qingju, B.~Chiwu, L.~Yuanlin, Q.~Litao, and Y.~Zongyan, ``A new signal
  processing algorithm of pulsed infrared thermography,'' \emph{Infrared
  Physics \& Technology}, vol.~68, pp. 173 –-- 178, 2015.

\bibitem{vrabie2012active}
V.~Vrabie, E.~Perrin, J.~Bodnar, K.~Mouhoubi, and V.~Detalle, ``Active ir
  thermography processing based on higher order statistics for nondestructive
  evaluation,'' in \emph{Signal Processing Conference (EUSIPCO), 2012
  Proceedings of the 20th European}.\hskip 1em plus 0.5em minus 0.4em\relax
  IEEE, 2012, pp. 894--898.

\bibitem{Zhang2001a}
J.~Q. Zhang and Y.~Yan, ``A wavelet-based approach to abrupt fault detection
  and diagnosis of sensors,'' \emph{Instrumentation and Measurement, IEEE
  Transactions on}, vol.~50, no.~5, pp. 1389--1396, 2001.

\bibitem{Ece2004}
D.~G. Ece and O.~N. Gerek, ``Power quality event detection using joint
  2-d-wavelet subspaces,'' \emph{Instrumentation and Measurement, IEEE
  Transactions on}, vol.~53, no.~4, pp. 1040--1046, 2004.

\bibitem{Blanco-Velasco2008}
M.~Blanco-Velasco, B.~Weng, and K.~E. Barner, ``Ecg signal denoising and
  baseline wander correction based on the empirical mode decomposition,''
  \emph{Computers in biology and medicine}, vol.~38, no.~1, pp. 1--13, 2008.

\bibitem{Cusido2008}
J.~Cusido, L.~Romeral, J.~A. Ortega, J.~A. Rosero, and A.~Garcia~Espinosa,
  ``Fault detection in induction machines using power spectral density in
  wavelet decomposition,'' \emph{Industrial Electronics, IEEE Transactions on},
  vol.~55, no.~2, pp. 633--643, 2008.

\bibitem{khan2008source}
A.~A. Khan, V.~Vrabie, J.~I. Mars, A.~Girard, and G.~d'Urso, ``A source
  separation technique for processing of thermometric data from fiber-optic dts
  measurements for water leakage identification in dikes,'' \emph{Sensors
  Journal, IEEE}, vol.~8, no.~7, pp. 1118--1129, 2008.

\bibitem{Purushotham2005}
V.~Purushotham, S.~Narayanan, and S.~A. Prasad, ``Multi-fault diagnosis of
  rolling bearing elements using wavelet analysis and hidden markov model based
  fault recognition,'' \emph{Ndt \& E International}, vol.~38, no.~8, pp.
  654--664, 2005.

\bibitem{Youssef2003}
O.~A. Youssef, ``A wavelet-based technique for discrimination between faults
  and magnetizing inrush currents in transformers,'' \emph{Power Delivery, IEEE
  Transactions on}, vol.~18, no.~1, pp. 170--176, 2003.

\bibitem{Zhang2001}
J.~Q. Zhang and Y.~Yan, ``A wavelet-based approach to abrupt fault detection
  and diagnosis of sensors,'' \emph{Instrumentation and Measurement, IEEE
  Transactions on}, vol.~50, no.~5, pp. 1389--1396, 2001.

\bibitem{Liu2008}
J.~Liu, W.~Wang, and F.~Golnaraghi, ``An extended wavelet spectrum for bearing
  fault diagnostics,'' \emph{Instrumentation and Measurement, IEEE Transactions
  on}, vol.~57, no.~12, pp. 2801--2812, 2008.

\bibitem{russakoff2004image}
D.~B. Russakoff, C.~Tomasi, T.~Rohlfing, and C.~R. Maurer~Jr, ``Image
  similarity using mutual information of regions,'' in \emph{Computer
  Vision-ECCV 2004}.\hskip 1em plus 0.5em minus 0.4em\relax Springer, 2004, pp.
  596--607.

\bibitem{Pluim2003}
J.~P. Pluim, J.~A. Maintz, and M.~A. Viergever, ``Mutual-information-based
  registration of medical images: a survey,'' \emph{Medical Imaging, IEEE
  Transactions on}, vol.~22, no.~8, pp. 986--1004, 2003.

\bibitem{Crum2014}
W.~R. Crum, T.~Hartkens, and D.~Hill, ``Non-rigid image registration: theory
  and practice,'' \emph{The British Journal of Radiology}, vol.~77, no.~2, pp.
  S140--53, 2004.

\bibitem{Hirschmuller2008}
H.~Hirschmuller, ``Stereo processing by semiglobal matching and mutual
  information,'' \emph{Pattern Analysis and Machine Intelligence, IEEE
  Transactions on}, vol.~30, no.~2, pp. 328--341, 2008.

\bibitem{Peng2005}
H.~Peng, F.~Long, and C.~Ding, ``Feature selection based on mutual information
  criteria of max-dependency, max-relevance, and min-redundancy,''
  \emph{Pattern Analysis and Machine Intelligence, IEEE Transactions on},
  vol.~27, no.~8, pp. 1226--1238, 2005.

\bibitem{corsini2009image}
M.~Corsini, M.~Dellepiane, F.~Ponchio, and R.~Scopigno, ``Image-to-geometry
  registration: a mutual information method exploiting illumination-related
  geometric properties,'' in \emph{Computer Graphics Forum}, vol.~28,
  no.~7.\hskip 1em plus 0.5em minus 0.4em\relax Wiley Online Library, 2009, pp.
  1755--1764.

\bibitem{Gueguen2014}
L.~Gueguen, S.~Velasco-Forero, and P.~Soille, ``Local mutual information for
  dissimilarity-based image segmentation,'' \emph{Journal of mathematical
  imaging and vision}, vol.~48, no.~3, pp. 625--644, 2014.

\end{thebibliography}
\end{document}